%% file: nas.tex
\documentclass{article} 
\usepackage{iclr2022_conference,times}

\input{math_commands.tex}

\usepackage{enumitem}
\usepackage{hyperref}

\usepackage{url}

\usepackage{graphicx} 
\usepackage{multirow} 
\usepackage{multicol} 
\usepackage{booktabs} 
\usepackage{arydshln} 
\usepackage{float}
\restylefloat{table}

\definecolor{amethyst}{rgb}{0.6, 0.4, 0.8} 
\definecolor{brown(web)}{rgb}{0.65, 0.16, 0.16} 
\usepackage{xcolor}

\usepackage{pifont}
\newcommand{\cmark}{\ding{51}}
\newcommand{\xmark}{\ding{55}}
\iclrfinalcopy

\usepackage{color, colortbl}

\usepackage{minitoc}

\definecolor{LightCyan}{rgb}{0.88,1,1}

\title{Natural Attribute-based Shift Detection}

\author{Jeonghoon Park$^{1}$\thanks{Equal contribution} , Jimin Hong$^{1}$\footnotemark[1] , Radhika Dua$^{1}$\footnotemark[1] , Daehoon Gwak$^{1}$,Yixuan Li$^{2}$, \\
 \textbf{Jaegul Choo$^{1}$, Edward Choi$^{1}$} \\
 KAIST$^{1}$, University of Wisconsin-Madison$^2$ \\
 \texttt{\{jeonghoon\_park, jimmyh, radhikadua, daehoon.gwak, jchoo,}\\ \texttt{edwardchoi\}@kaist.ac.kr}, \\
\texttt{sharonli@cs.wisc.edu}
}

\begin{document}

\maketitle

\begin{abstract}

Despite the impressive performance of deep networks in vision, language, and healthcare,
unpredictable behaviors on samples from the distribution different than the training distribution cause severe problems in deployment.
For better reliability of neural-network-based classifiers, we define a new task, natural attribute-based shift (NAS) detection, to detect the samples shifted from the training distribution by some natural attribute such as age of subjects or brightness of images.
Using the natural attributes present in existing datasets, we introduce benchmark datasets in vision, language, and medical for NAS detection.
Further, we conduct an extensive evaluation of prior representative out-of-distribution (OOD) detection methods on NAS datasets and observe an inconsistency in their performance.
To understand this, we provide an analysis on the relationship between the location of NAS samples in the feature space and the performance of distance- and confidence-based OOD detection methods.
Based on the analysis, we split NAS samples into three categories and further suggest a simple modification to the training objective to obtain an improved OOD detection method that is capable of detecting samples from all NAS categories.
\end{abstract}

\section{Introduction}
\vspace{-3mm}
\input{010introduction}

\section{Natural Attribute-based Shift Detection}
\vspace{-3mm}
\input{020problem_setup}

\section{NAS Dataset Description}
\vspace{-3mm}
\input{021dataset}

\section{Can OOD Detection Methods also Detect NAS?}
\vspace{-3mm}
\input{030initial_experiments}

\vspace{-3mm}
\section{Analyzing Inconsistency of OOD Detection Methods}
\vspace{-3mm}
\label{sec:initial_exp}
\input{040discussion}

\section{Method for Consistent NAS Detection Performance}
\vspace{-3mm}
\input{050new_method}

\section{Related Work}
\vspace{-3mm}
\input{060relatedwork}

\section{Conclusions}
\vspace{-3mm}
\input{070conclusions}

\bibliography{nas.bbl}
\bibliographystyle{iclr2022_conference}

\newpage
\clearpage

\input{main_appendix}

\end{document}

%% file: math_commands.tex
\usepackage{amsmath,amsfonts,bm}

\def\eqref#1{equation~\ref{#1}}

\def\1{\bm{1}}

\DeclareMathAlphabet{\mathsfit}{\encodingdefault}{\sfdefault}{m}{sl}
\SetMathAlphabet{\mathsfit}{bold}{\encodingdefault}{\sfdefault}{bx}{n}



%% file: 010introduction.tex
Deep learning has significantly improved the performance in
various domains such as computer vision, natural language processing, and healthcare~\citep{bojarski2016end, mikolov2013efficient, dermato16deep}.
However, it has been reported that the deep classifiers make unreliable predictions on samples drawn from a different distribution than the training distribution~\citep{AmodeiOSCSM16, hendrycks2017base, NguyenYC15}.
Especially, this problem can become severe when the test distribution is shifted from the training distribution by some attributes (\textit{i.e.}, age of subjects or brightness of images),
as such shift could gradually degrade the classifier performance until its malfunction is explicitly visible.
These shifts occur in the real-world as a result of a change in specific attribute.
For example, a clinical text-based diagnosis classifier trained in $2021$ will gradually encounter increasingly shifted samples as time flows, since writing styles change and new terms are introduced in time.
Detection of such samples is a vital task especially in safety-critical systems, such as autonomous vehicle control or medical diagnosis, where wrong predictions can lead to dire consequences.
To this end, we take a step forward by proposing a new task of detecting samples shifted by a natural attribute (\textit{e.g.}, age, time) that can easily be observed in the real-world setting.
We refer to such shifts as \textit{\textbf{N}atural \textbf{A}ttribute-based \textbf{S}hifts (NAS)}, and the task of detecting them as NAS detection.

Detection of NAS is both different from, and also more challenging than out-of-distribution (OOD) detection~\citep{hendrycks2017base,liang2017enhancing,mahala18,sastry2021gram}, which typically evaluates the detection methods with a clearly distinguished in-distribution (ID) samples and OOD samples (\textit{e.g.}, CIFAR10 as ID and SVHN as OOD, which have disjoint labels).
In contrast, we aim to detect samples from a natural attribute-based shift within the same label space.
Since NAS samples share more features with the ID than the typical OOD samples do, identifying the former is expected to be more challenging than the latter.
Although OOD detection has some relevance to NAS detection, comprehensive evaluation of the existing OOD detection methods on the natural attribute-based shift is an unexplored territory.
Therefore, in this paper, we perform an extensive evaluation of representative OOD methods on NAS samples.

Depending on the task environment, NAS detection can be pursued in parallel to domain generalization~\citep{gen1, gen2, gen3}, which aims to overcome domain shifts (\textit{e.g.}, image classifier adapting to sketches, photos, art paintings, etc.).
For example, an X-ray-based diagnosis model should detect images of unusual brightness so that the X-ray machine can be properly configured, and the diagnosis model can perform in the optimal setting.
In other cases, domain generalization can be preferred, such as when we expect the classifier to be deployed in a less controlled environment (\textit{e.g.}, online image classifier) for non-safety critical tasks.

In this paper, we formalize NAS detection to enhance the reliability of real-world decision systems.
Since there exists no standard dataset for this task, we create a new benchmark dataset in the vision, text, and medical domain by adjusting the natural attributes (\textit{e.g.}, age, time, and brightness) of the ID dataset.
Then we conduct an extensive evaluation of representative confidence- and distance-based OOD methods on our datasets and observe that none of the methods perform consistently across all NAS datasets.

After a careful analysis on where NAS samples reside in the feature space and its impact on the distance- and confidence-based OOD detection performance, we identify the root cause of the inconsistent performance.
Following this observation, we define three general NAS categories based on two criteria: the distance between NAS samples and the decision boundary, the distance between NAS samples, and the ID data.
Finally, we conduct an additional experiment to demonstrate that a simple modification to the negative log-likelihood training objective can dramatically help the Mahalanobis detector~\citep{mahala18}, a distance-based OOD detection method, generalize to all NAS categories. 
We also compare our results with various baselines and show that our proposed modification outperforms the baselines and is effective across the three NAS datasets.

In summary, the contributions of this paper are as follows:
\vspace{-2mm}
\begin{itemize}[leftmargin=5.5mm]
    \setlength\itemsep{0.2em}
    \item We define a new task, \textit{\textbf{N}atural \textbf{A}ttribute-based \textbf{S}hift detection (NAS detection)}, which aims to detect the samples from a distribution shifted by some natural attribute. We create a new benchmark dataset and provide them to encourage further research on evaluating NAS detection.
    \item To the best of our knowledge, this is the first work to conduct a comprehensive evaluation of the OOD detection methods on shifts based on natural attributes, and discover that none of the OOD methods perform consistently across all NAS scenarios.
    \item We provide novel analysis based on the location of shifted samples in the feature space and the performance of existing OOD detection methods. Based on the analysis, we split NAS samples into three categories.
    \item We demonstrate that a simple yet effective modification to the training objective for deep classifiers enables consistent OOD detection performances for all NAS categories. 
\end{itemize}

%% file: 020problem_setup.tex
We now formalize a new task, NAS detection, which aims to enhance the reliability of real-world decision systems by detecting samples from NAS.
We address this task in the classification problems. 
Let $\mathcal{D}_{\text{I}} =\{ \mathcal{X}, \mathcal{Y} \}$ denote the in-distribution data, which is composed of $N$ training samples with inputs $\mathcal{X}=\{x_1, ..., x_N\}$ and labels $\mathcal{Y}=\{y_1, ..., y_N\}$.
Specifically, $x_i\in \mathbb{R}^{d}$ represents a $d$-dimensional input vector, and $y_i \in \mathcal{K} $ represents its corresponding label where $\mathcal{K}=\{1,...,K\}$ is a set of class labels.
The discriminative model $f_{\theta}: \mathcal{X} \rightarrow \mathcal{Y}$ learns with ID dataset $\mathcal{D}_{\text{I}}$ to assign label $y_i$ for each $x_i$. 
In the NAS detection setting, we assume that an in-distribution sample consists of attributes, and some of the attributes can be shifted in the test time due to natural causes such as time, age, or brightness.

When a particular attribute $A$ ($\textit{e.g.}$, age), which has a value of $a$ ($\textit{e.g.}$, 16), is shifted by the degree of $\delta$, the shifted distribution can be denoted as $\mathcal{D}^{A = a+\delta}_{\text{S}} =\{ \mathcal{X}^\prime, \mathcal{Y}^\prime \}$.
$\mathcal{X^\prime}=\{x_1^\prime, ..., x_M^\prime\}$ and $\mathcal{Y^\prime}=\{y_1^\prime, ..., y_M^\prime\}$ represents the $M$ shifted samples and labels respectively. 
Importantly, in the NAS setting, although the test distribution is changed from the ID, the label space is preserved as $\mathcal{K}$, which is the set of class labels in $\mathcal{D}_{\text{I}}$. 
In the test time, the model $f_{\theta}$ might encounter the sample $x^{\prime}$ from a shifted data $\mathcal{D}^{A = a+\delta}_{\text{S}}$, and it should be able to identify that the attribute-shifted sample is not from the ID.

%% file: 021dataset.tex
\label{sec:dataset}

In this section, we describe three benchmark datasets which have a controllable attribute for simulating realistic distribution shifts.
Since there exists no standard dataset for NAS detection, we create new benchmark datasets using existing datasets by adjusting natural attributes in order to reflect real-world scenarios. 
We carefully select datasets from vision, language, and medical domains containing natural attributes
(\textit{e.g.}, year, age, and brightness), which allows us to naturally split the samples.
By grouping samples based on these attributes, we can induce natural attribute-based distribution shifts as described below.

\begin{figure}[t]
\vspace{-10mm}
\centering
     \includegraphics[width=1.0\textwidth]{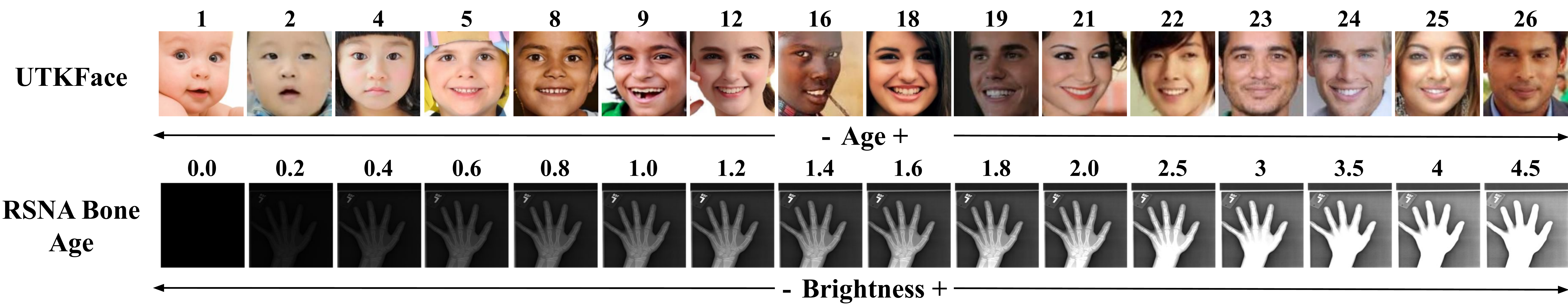}
     \vspace{-7mm}
     \caption{ Facial images from the UTKFace dataset to show the variation with age. X-ray images with different levels of brightness created from the RSNA Bone Age dataset.}
     \label{fig:dataset}
\vspace{-3mm}
\end{figure}

\textbf{Image. \enskip} We use the UTKFace dataset~\citep{zhifei2017cvpr} which consists of over $20,000$ face images with annotations of age, gender, and ethnicity. As shown in Figure~\ref{fig:dataset}, we can visually observe that the facial images vary with age. 
Therefore, we set the $1,282$ facial images of $26$ years old age as $\mathcal{D}_{\text{I}}$.
For creating the NAS dataset, we vary the age of UTKFace dataset. To obtain an equal number of samples in the NAS dataset, the age groups that has less than $200$ images are merged into one group until it has $200$ samples. Finally, $15$ groups $\mathcal{D}_{\text{S}}^{\text{age}}$ are produced for the NAS datasets, varying the ages from $25$ to $1$ (\textit{i.e.}, $\mathcal{D}_{\text{S}}^{\text{age}=25}, \mathcal{D}_{\text{S}}^{\text{age}=24},\ldots, \mathcal{D}_{\text{S}}^{\text{age}={1}}$).

\textbf{Text. \enskip} We use the Amazon Review dataset~\citep{HeM16, McAuleyTSH15} which contains product reviews from Amazon. We consider the product category "book" and group its reviews based on the year to reflect the distributional shift across time.
We obtain $9$ groups with each group containing reviews from the year between $2005$ and $2014$. Then, the group with $24,000$ reviews posted in $2005$ is set as $\mathcal{D}_{\text{I}}$, and the groups with reviews after $2005$ as $\mathcal{D}_{\text{S}}^{\text{year}}$ (\textit{i.e.}, $\mathcal{D}_{\text{S}}^{\text{year}=2006}, \mathcal{D}_{\text{S}}^{\text{year}=2007},\ldots, \mathcal{D}_{\text{S}}^{\text{year}={2014}}$). Each $\mathcal{D}_{\text{S}}^{\text{year}}$ group contains $1500$ positive reviews and $1500$ negative reviews. We observed that as we move ahead in time, the average length of a review gets shorter and it uses more adjectives than previous years. Due to the space constraint, we provide a detailed analysis of the dataset in the Section~\ref{sec:supp_dataset_details} of the Appendix. 

\textbf{Medical. \enskip} We use the RSNA Bone Age dataset~\citep{013a4b4eeca44ecab48a66956acbb91d}, a real-world dataset that contains left-hand X-ray images of the patient, along with their gender and age ($0$ to $20$ years). 
We consider patients in the age group of $10$ to $12$ years for our dataset.
To reflect diverse X-ray imaging set-ups in the hospital, 
we varied the brightness factor between 0 and 4.5 and form 16 different dataset $\mathcal{D}_{\text{S}}^{\text{brightness}}$ (\textit{i.e.}, $\mathcal{D}_{\text{S}}^{\text{brightness}=0.0}, \mathcal{D}_{\text{S}}^{\text{brightness}=0.2},\ldots, \mathcal{D}_{\text{S}}^{\text{brightness}={4.5}}$), and each group contains X-ray images of $200$ males and $200$ females.
Figure~\ref{fig:dataset} presents X-ray images with different levels of brightness with realistic and continuous distribution shifts.
In-distribution data $\mathcal{D}_{\text{I}}$ is composed of $3,000$ images of brightness factor $1.0$ (unmodified images).

%% file: 030initial_experiments.tex
In this section, we briefly discuss about OOD detection methods and conduct an extensive evaluation of OOD detection methods on our proposed benchmark datasets.

\vspace{-2mm}
\subsection{OOD Detection Methods}
\vspace{-2mm}

In this work, we use three widely-used post-hoc and modality-agnostic OOD detection methods. We use \textbf{maximum of softmax probability (MSP)}~\citep{hendrycks2017base} and \textbf{ODIN}~\citep{liang2017enhancing} as confidence-based OOD detection baselines, and \textbf{Mahalanobis detector}~\citep{mahala18} as distance-based OOD detection baseline.
Note that \textbf{ODIN} and \textbf{Mahalanobis detector} assume the availability of OOD validation dataset to tune their hyperparameters.
However, for all our experiments, we use variants of the above methods that do not access the OOD validation dataset as \cite{godin}.
The exact equations and details of how each OOD detection method assigns an OOD score to a given sample is provided in Section~\ref{sec:supp_ood_details} of the Appendix.

\vspace{-2mm}
\subsection{Experiments and Results}
\vspace{-2mm}
\label{sec:initial_exp_result}
We now systematically evaluate the performance of the three OOD detection methods under NAS.
We report the AUROC of all OOD detection methods averaged across five random seeds, evaluated for all NAS datasets.

\textbf{Experimental Settings. \enskip} 
In image domain, we train a gender classification model on our UTKFace NAS dataset using ResNet18 model and the cross-entropy loss.
We use our Amazon Book Review NAS dataset in text domain and train a 4-layer Transformer with the cross-entropy loss for the sentiment classification task.
Lastly, in medical domain, we use our RSNA Bone Age NAS dataset and train a ResNet18 with cross-entropy loss to predict the gender given the hand X-ray image of the patient. We then evaluate the trained model on the corresponding test set in image, text, medical domains, respectively.
Further, we evaluate the NAS detection performance of representative OOD detection methods in image, text, and medical domains on their corresponding NAS datasets, which gradually shift with age, year, and brightness, respectively.

\begin{figure}[t]
\vspace{-10mm}
\centering
     \includegraphics[width=\textwidth]{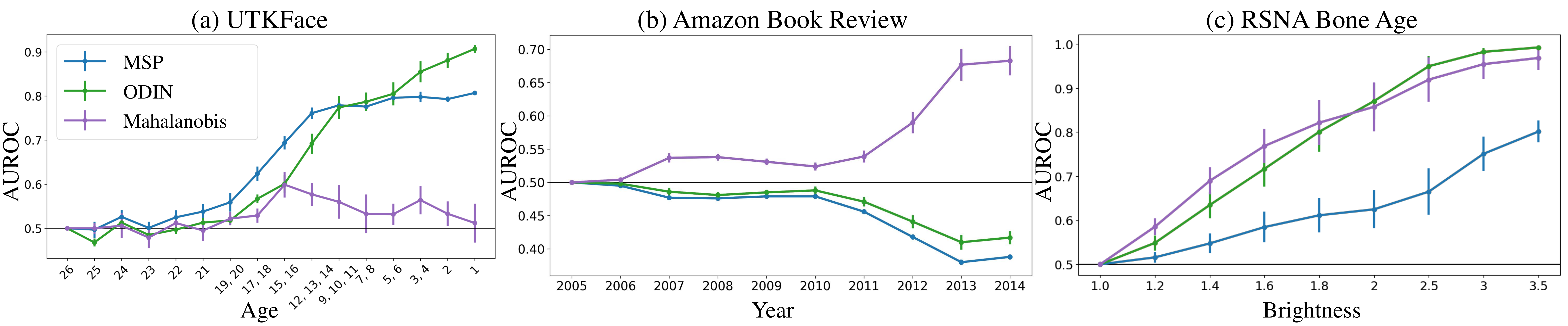}
\vspace{-7mm}
\caption{Comparison of well-known distance-based and confidence-based OOD detection methods for the CE model on our benchmark datasets. Age '26', year '2005', and brightness '1.0' are in-distribution data in UTKFace, Amazon Review, and RSNA Bone Age dataset, respectively. The NAS detection performance of these methods is inconsistent across different datasets.
}\label{fig:init_exp_results}
\vspace{-3mm}
\end{figure}

\textbf{Results. \enskip} 
We present the classification accuracy of the trained models on the ID test set in Table~\ref{tab:accuracy}. We observe that the models trained using the cross-entropy loss obtain high accuracy and perform well on their corresponding tasks.
We further demonstrate the effectiveness of the existing representative OOD detection methods on our benchmark datasets in Figure~\ref{fig:init_exp_results}. We observe that
in the UTKFace NAS dataset, samples are detected by ODIN and MSP, which are confidence-based methods, but not by Mahalanobis detector (Figure~\ref{fig:init_exp_results}a).
In the Amazon Review dataset, NAS samples are detected only by the Mahalanobis detector, while MSP and ODIN fail (Figure~\ref{fig:init_exp_results}b). Moreover, the scores of confidence-based methods are lower than 50 in AUROC.
Lastly, Figure~\ref{fig:init_exp_results}c shows that inputs from NAS in the RSNA Bone Age dataset are detected well by all three methods.

%% file: 040discussion.tex
In this section, we first study the behavior of NAS samples in three datasets using PCA visualization.
Then we analyze the inconsistent performance of the OOD detection methods by considering them in two categories, namely confidence-based and distance-based methods.
Lastly, based on the analysis, we conclude this section by defining three NAS categories.

\vspace{-2mm}
\subsection{Analysis of the location of NAS samples} 
\begin{figure}[t]
\vspace{-7mm}
\centering
     \includegraphics[width=1.0\textwidth]{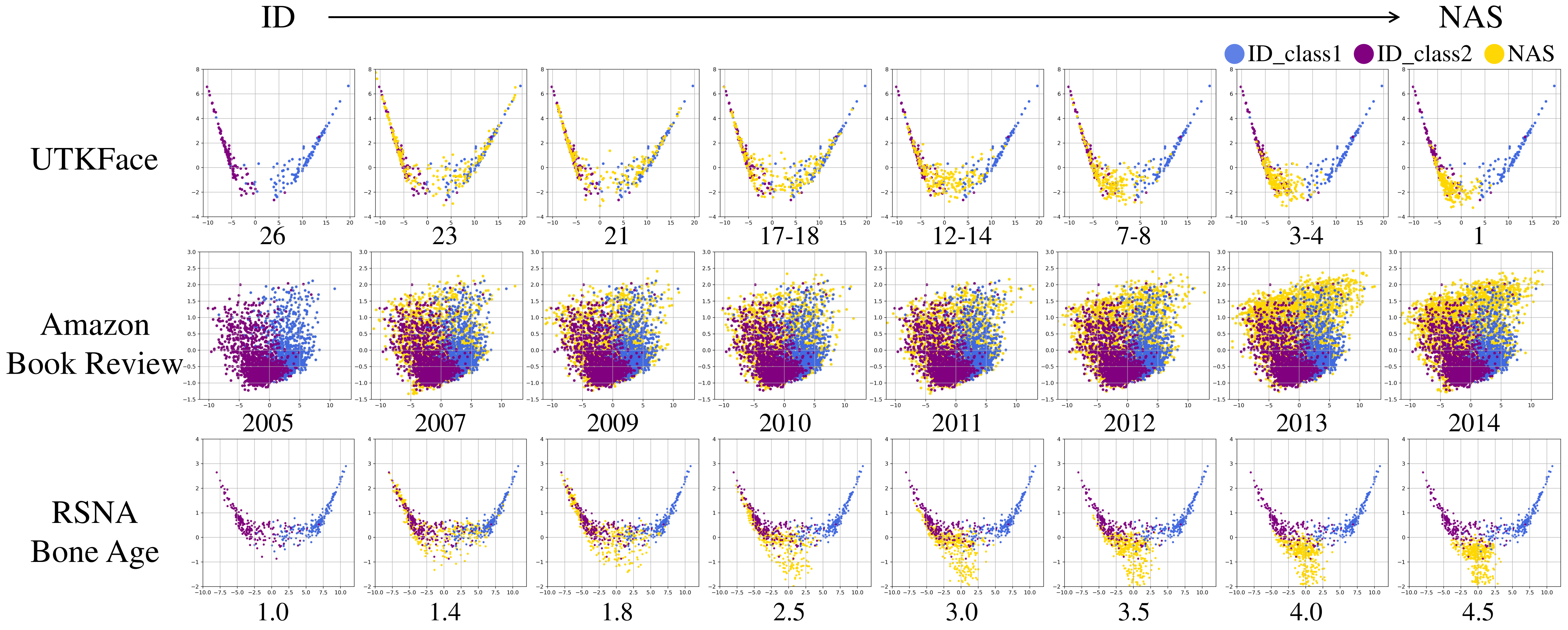}
    \vspace{-7mm}
    \caption{PCA visualization to demonstrate the movement of NAS samples as we vary the age, year and brightness in UTKFace, Amazon Review and RSNA Bone Age dataset respectively.}
    \label{fig:pca}
\vspace{-2mm}
\end{figure}

\begin{figure}[t]
    \centering
    \includegraphics[width=1.0\textwidth]{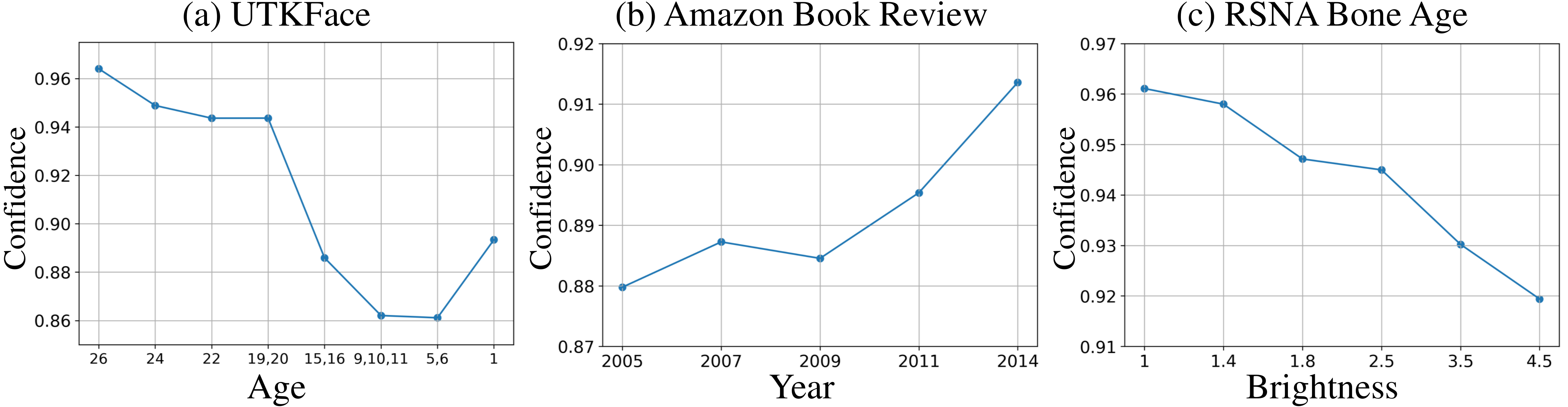}
    \vspace{-6mm}
    \caption{Impact on prediction confidence on varying age, year, and brightness in UTKFace, Amazon Review, and RSNA Bone Age dataset, respectively.}
    \label{fig:acc_conf}
\vspace{-5mm}
\end{figure}

\vspace{-2mm}
\label{sec:analaysis_of_inconsistency}
As illustrated in Figure~\ref{fig:pca}, we apply principal component analysis (PCA) on the feature representations obtained from the penultimate layer of the models to visualize the movement of NAS samples as we monotonically increase the degree of attribute shift (\textit{i.e.}, age, year, and brightness).
Further, Figure~\ref{fig:acc_conf} presents the model's prediction confidence across varying degrees of the attribute shift.

\textbf{Image. \enskip} By gradually changing the age, NAS samples move toward the space between the two clusters of ID samples (\textit{i.e.}, the decision boundary) as can be seen in Figure~\ref{fig:pca} [Top].
Further, Figure ~\ref{fig:acc_conf}a demonstrates that confidence decrease as we increase the degree of attribute shift, which indicates that NAS samples move close to the decision boundary.
Note that the majority of the NAS samples still overlap with ID sample clusters as we change the age.

\textbf{Text. \enskip} As shown in Figure~\ref{fig:pca} [Middle], NAS samples gradually move away from the ID samples (and away from the decision boundary) as the year changes.
In contrast to the UTKFace dataset, the confidence gradually increases, as shown in Figure~\ref{fig:acc_conf}b, since the NAS samples are getting far away from the decision boundary.

\textbf{Medical. \enskip} Figure~\ref{fig:pca} [Bottom] demonstrates that when we increase the brightness, NAS samples move to the middle of the two classes and also move towards the outer edge of the ID sample clusters.
Furthermore, as shown in Figure~\ref{fig:acc_conf}c, the relatively decreased confidence indicates that NAS samples are placed near the decision boundary as we increase the brightness of the images.

\vspace{-2mm}
\subsection{Comparison between Confidence-based and Distance-based OOD Detection}
\label{sec:comp_conf_dist}
\vspace{-2mm}

\textbf{Confidence-based methods. \enskip} Figure~\ref{fig:init_exp_results} and Figure~\ref{fig:pca} illustrate that confidence-based methods achieve high AUROC when the NAS samples are near the decision boundary due to their low confidence.
In the image and medical domain, we observed the NAS samples moving towards the decision boundary with increasing attribute shift, thus they are detected by the confidence-based methods.
In contrast, in the text domain, the high prediction confidence of the shifted NAS samples causes the degradation in the AUROC of the confidence-based methods.
Therefore, we conclude that to effectively utilize the confidence-based methods in all three NAS datasets, it is necessary to reduce the confidence of samples outside the ID, namely, enforce NAS samples to move near the decision boundary, which is not always possible (\textit{e.g.,} Amazon Review dataset).

\textbf{Distance-based methods. \enskip} From Figure~\ref{fig:init_exp_results} and Figure~\ref{fig:pca}, we observe that the distance-based OOD detection method (\textit{i.e.}, Mahalanobis Detector) achieves high AUROC when NAS samples are sufficiently away from the ID samples. 
In the text and medical domains, the Mahalanobis detector worked well since NAS samples moved sufficiently away from the ID samples as the shift increased.
However, in the image domain, the method fails to detect NAS samples because instead of deviating from the ID, they move intermediately between the classes. 

Prior works~\citep{gsoftmax,liu2017sphereface} report that the cross-entropy loss cannot guarantee a sufficient inter-class distance.
In other words, representations do not need to be far from the decision boundary to lower the cross-entropy loss.
In this regard, we assume that the performance degradation of the Mahalanobis detector is caused by the cross-entropy loss learning latent features that are not separable enough to detect the NAS samples located between the classes (\textit{e.g.}, Figure~\ref{fig:pca} [Top]).
Specifically, if some classes are located nearby in the feature space, samples moving between classes (\textit{i.e.}, the case of the UTKFace dataset) will not be far from the ID.
Even though NAS samples move away from one of the ID class cluster, they will gradually get closer to another ID class cluster.

\vspace{-2mm}
\subsection{NAS Categorization}
\vspace{-2mm}

\begin{figure}[ht!]
\vspace{-1mm}
\centering
    \includegraphics[width=\textwidth]{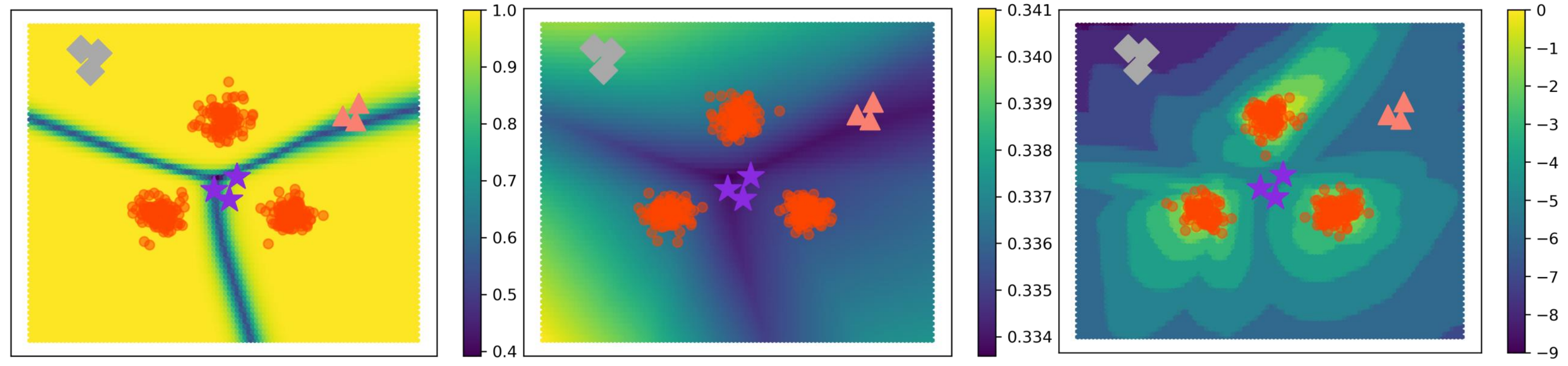}
    \vspace{-7mm}
    \caption{\small ID score landscape (brighter region means higher ID score) of the existing OOD detection methods (left: MSP, middle: ODIN, right: Mahalanobis). We use a synthetic 2D dataset to train a 4-layer ResNet. The \textbf{\textcolor{red}{Red points}} represent the ID samples; \textbf{\textcolor{amethyst}{Purple Stars}}, \textbf{\textcolor{darkgray}{Gray Diamonds}} and \textbf{\textcolor{orange}{Orange Triangles}} indicate samples from different NAS categories. A sample is regarded as NAS when it has a low ID score.}
    \label{fig:synthetic_ood}
\end{figure}
\input{tabels/categories}

Considering the performance of confidence-based and distance-based detection methods, we now divided NAS into three categories based on two criteria: 1) whether the dataset is near the decision boundary or not; 2) whether they are close or far from the in-distribution dataset. Since a dataset (\textit{i.e.}, samples) far from the decision boundary and overlapping with the in-distribution dataset is not a distributionally shifted dataset, our work focuses on the remaining three cases that cover all possible scenarios of NAS.
Without loss of generality, we use a classification task with three classes as a motivating example depicted by Figure~\ref{fig:synthetic_ood}.

\vspace{-2mm}
\begin{itemize}[leftmargin=3mm]
\setlength\itemsep{-0.5em}
    \item \textbf{NAS category 1:} This category comprises of NAS samples that are located near the decision boundary, and between ID samples of different classes. \textbf{\textcolor{amethyst}{Purple Stars}} in Figure~\ref{fig:synthetic_ood} represents samples from this category.
    Such samples are easily detected by confidence-based methods, MSP and ODIN, but harder to be detected by Mahalanobis Detector, which is a distance-based method.\\
    
    \vspace{-2mm}
    \item \textbf{NAS category 2:} This category consists of NAS samples that are placed away from the decision boundary and the ID data. For example, \textbf{\textcolor{darkgray}{Gray Diamonds}} in Figure~\ref{fig:synthetic_ood}. Mahalanobis detector regards such samples as NAS, whereas confidence-based methods fail to detect them since NAS samples have a higher prediction confidence (\textit{i.e.}, higher ID score) than ID samples.\\
    
    \vspace{-2mm}
    \item \textbf{NAS category 3:}  This category mainly comprises of NAS samples located anywhere near the decision boundary but far away from ID data. For example, \textbf{\textcolor{orange}{Orange Triangles}} in Figure~\ref{fig:synthetic_ood}. 
    Such samples are easily detected by both distance-based and confidence-based OOD detection methods.
\end{itemize}

%% file: tabels/categories.tex
\begin{table}[ht!]
\footnotesize
\begin{center}
\begin{tabular}{ c c c c } 
  \toprule
  & Category 1 & Category 2 & Category 3 \\ 
  \midrule
  Decision Boundary & Near & Far & Near \\
  In-distribution & Near & Far & Far \\
  Confidence-based methods works? & \cmark & \xmark & \cmark \\
 Distance-based methods works? & \xmark & \cmark & \cmark \\
  \toprule
\end{tabular}
\vspace{-5mm}
\caption{Comparison between different categories of NAS based on the location of samples in the features space and performance of confidence-based and distance-based methods.}
\label{tab:accuracy}
\end{center}
\end{table}

%% file: 050new_method.tex
In this section, we suggest a modification in the training objective for deep classifiers to encourage consistent NAS detection performance on all NAS categories.
Then we provide experiment results where the proposed method was compared against diverse OOD detection methods on the three NAS datasets (UTKFace, Amazon Review, RSNA Bone Age).

\vspace{-2mm}
\subsection{Method}
\vspace{-2mm}
\label{sec:new_method}
For a generally applicable OOD detection method to all NAS categories, we suggest a new training objective for deep classifiers comprised of \textit{classification loss} ($\mathcal{L}_\text{CE}$), \textit{distance loss} ($\mathcal{L}_\text{dist}$)  and \textit{entropy loss} ($\mathcal{L}_\text{entropy})$.
The proposed objective improves the performance of the Mahalanobis detector on NAS samples from category $1$ without sacrificing performance on NAS samples from other categories.
We focus on improving the distance-based OOD detection method rather than the confidence-based method since it is not always possible to enforce NAS samples to be embedded near the decision boundary, as discussed in Section~\ref{sec:comp_conf_dist}.

The proposed training loss is defined as: 
\vspace{-1mm}
\begin{equation}
    \mathcal{L}_\text{total} = \mathcal{L}_\text{CE} + \mathcal{L}_\text{dist} + \mathcal{L}_\text{entropy}.
    \label{eq:total_loss}
\end{equation}
Note that for classification loss, we use the standard cross-entropy loss, but other losses such as focal loss can also be used.
The distance loss is used to increase the distance between distinct class distributions of ID samples so that NAS samples have larger space to move around without overlapping with ID clusters, especially in NAS category $1$:
\vspace{-1mm}
\begin{equation}
    \mathcal{L}_\text{dist} = \lambda_{1} \frac{1}{\binom{K}{2}}\sum_{l}\sum_{k\neq l}\frac{\lVert \mu_l - \mu_k \rVert_2}{\sqrt{D}},
\end{equation}
where $\lambda_1 < 0$ is a hyperparameter, $K$ is the number of target classes, $D$ is the dimension of feature representation in the latent space (typically the penultimate layer), and $\mu_l$ is the mean vector of the features of samples with label $l$. Since the value of the vector norm increases as the dimension of the feature space increases, we normalize the distance by the square root of the feature dimension.

As will be discussed in Section~\ref{sec:supp_ablations} of Appendix, we discovered after some initial experiments that the \textit{distance loss} often made the model use a very limited number of latent dimensions to increase the distance between class mean vectors, which degraded the NAS detection performance.
In other words, adding only $\mathcal{L}_\text{dist}$ to $\mathcal{L}_\text{CE}$ caused the latent feature space to collapse into a very small number of dimensions (\textit{i.e.} rank-deficient), which caused all NAS samples to be embedded near the ID samples.
Therefore, we add $\textit{entropy loss}$ to increase the number of features used to represent samples.

The $\textit{entropy loss}$ is defined as:
\vspace{-1mm}
\begin{equation}
    \mathcal{L}_\text{entropy} = \lambda_{2}\frac{D}{\sum_{i} \mathrm{Var}(\mathbf{z}_i)} + \lambda_{3}\frac{1}{\binom{D}{2}}\sum_{i}\sum_{j \neq i} C_{ij}^2,
    \label{eq:entropy}
\end{equation}
where $\lambda_2>0$ and $\lambda_3>0$ are hyperparameters, $\mathbf{z} \in \mathbb{R}^{D}$ is the feature representation in the latent feature space, $\mathrm{Var}(\cdot)$ is variance, and $C_{ij}$ is the correlation coefficient between $i$th and $j$th dimension of the feature space.
Specifically, $C_{ij}$ is described as:
\vspace{-1mm}
\begin{equation}
    C_{ij} = \frac{\mathrm{Cov}\left(\mathbf{z}_i,\mathbf{z}_j\right)}{\sigma(\mathbf{z}_i)\sigma(\mathbf{z}_j)},
\end{equation}
where $\mathrm{Cov}(\cdot)$ and $\sigma(\cdot)$ are covariance and the standard deviation, respectively.

Intuitively, the first term in \eqref{eq:entropy} encourages each latent dimension to have diverse values, therefore preventing the latent feature space from collapsing into a confined space.
With the first term alone, however, all latent dimensions might learn correlated information, thus making the latent space rank-deficient.
Therefore, we use the second term in \eqref{eq:entropy} to minimize the correlation between different latent dimensions. Note that minimizing the feature correlation was also used in previous works under different contexts such as self-supervised learning~\citep{zbontar2021barlow}.

\input{tabels/accuracy}

\input{tabels/exp_table_posthoc}

\subsection{Results and Discussion}
\vspace{-2mm}

To demonstrate the effectiveness of the suggested method, we train all classifiers using the standard cross-entropy loss and our modified loss and compare post-hoc OOD detection methods across three NAS datasets.
Specifically, we present the results of the confidence-based (\textit{i.e.}, MSP and ODIN) and the distance-based methods (\textit{i.e.}, Mahalanobis distance).
We also include a recently proposed OOD detection method~\citep{sastry2021gram} which computes channel-wise correlations in CNN with the gram matrix and estimates the deviation of the test samples from the training samples to detect the OOD samples.
\footnote{However, in the text domain, the notion of gram matrix is vague, and hence we do not compare against this baseline.}
Although this method uses the distance in the channel correlation space, we expect it to behave more similarly to the Mahalanobis detector than confidence-based methods, namely MSP and ODIN.
We also compare with another recent baseline which exploits the energy score to detect OOD samples~\citep{liu2020energy}.
As the method leverage the logit layer to calculate the energy score, and the softmax score is based on the logit values, we conjecture that the energy score demonstrates detection ability similar to the confidence-based methods.

For a fair comparison, we use only the penultimate layer for evaluating the Mahalanobis detector and Gram matrix since our method is developed based on the analysis of the NAS samples in the penultimate layer feature space. 
We also provide results of the ensemble version that utilize all layers in Section~\ref{sec:supp_additional_baselines} of the Appendix.
We describe the details of experiment setups and selecting the values for $\lambda_1$, $\lambda_2$ and $\lambda_3$ in the Appendix Section~\ref{sec:supp_implementation_details}, where the three values can be reasonably chosen without any explicit NAS validation datasets.
We also present an ablation study to investigate the effect of different terms in the proposed loss in Section~\ref{sec:supp_ablations} of the Appendix.
We also compare with other recent baselines that suggest other training objective for OOD detection in Section~\ref{sec:supp_additional_baselines} of the Appendix.

As shown in Table~\ref{tab:accuracy}, the in-distribution classification accuracy of the model trained with our suggested loss is comparable to that of the model trained with the cross-entropy loss.
Further, we present the NAS detection performance of the baselines and our method in Table~\ref{tab:auroc}.
As we expected above, the gram matrix achieves performance similar to the Mahalanobis detector, demonstrating the low AUROC in the UTKFace dataset.
Interestingly, even though the energy score is calculated based on the logit values, which are more tempered than the softmax score, it shows NAS detection performance similar to that of MSP, which is a confidence-based method.
These results show that similar to MSP, ODIN, and Mahalanobis, Gram matrix and Energy-based methods also have performance inconsistency on NAS datasets.
We report the additional NAS detection performance on four other metrics, which are often used in the OOD detection community in Section~\ref{sec:supp_other_metrics} of Appendix.

Also, it is readily visible that our proposed training objective makes the Mahalanobis detector a robust NAS detection method for all NAS categories.
In UTKFace, we can see the dramatic NAS detection performance increase for Mahalanobis detector.
And for the other two datasets, the proposed loss does not decrease the NAS detection performance of Mahalanobis detector.
Note that ODIN is more sensitive to OOD samples than Mahalanobis detector for some brightness levels in the RSNA Bone Age dataset, but it shows inconsistent performance across all three datasets.

%% file: tabels/accuracy.tex
\begin{table*}
\vspace{-5mm}
\footnotesize
\begin{center}

\begin{tabular}{*{7}{c}}
\toprule
        &\multicolumn{2}{c}{\multirow{1}{*}{UTKFace}} 
        &\multicolumn{2}{c}{\multirow{1}{*}{Amazon Review}}
        &\multicolumn{2}{c}{\multirow{1}{*}{RSNA Bone Age}}
        \\ \cmidrule(lr){2-3} \cmidrule(lr){4-5} \cmidrule(lr){6-7} 
        & \multicolumn{1}{c}{CE} & \multicolumn{1}{c}{Ours}
        & \multicolumn{1}{c}{CE} & \multicolumn{1}{c}{Ours}
        & \multicolumn{1}{c}{CE} & \multicolumn{1}{c}{Ours}
        
        \\ \midrule
        
        Accuracy & 94.6{\scriptsize $\pm$0.6} & 94.1{\scriptsize $\pm$0.7} & 85.3{\scriptsize $\pm$0.9} & 84.0{\scriptsize $\pm$0.9} & 93.0{\scriptsize $\pm$0.6} & 92.3{\scriptsize $\pm$0.3} \\

\toprule

\end{tabular}

\caption{In-distribution classification accuracy on three datasets with cross-entropy loss and our proposed loss.}
\label{tab:accuracy}
\end{center}
\vspace{-5mm}
\end{table*}

%% file: tabels/exp_table_posthoc.tex
\begin{table*}[t]
\footnotesize
\centering
\vspace{-5mm}
\resizebox{\linewidth}{!}{
    \begin{tabular}{c c c  c  c  c c  c c }
        \toprule
        
        &\multicolumn{1}{c}{\multirow{2}{*}{Distribution}} & \multicolumn{1}{c}{\multirow{2}{*}{Age}} 
        & \multicolumn{5}{c}{\multirow{1}{*}{CE}}  & \multicolumn{1}{c}{Ours} \\ \cmidrule(lr){4-8}  \cmidrule(lr){9-9} 
        &\multicolumn{1}{c}{} & \multicolumn{1}{c}{} & \multicolumn{1}{c}{MSP} & \multicolumn{1}{c}{ODIN} & \multicolumn{1}{c}{Mahalanobis} & \multicolumn{1}{c}{Gram matrix} &  \multicolumn{1}{c}{Energy score} & \multicolumn{1}{c}{Mahalanobis}  
        
        \\ \midrule
        
        \parbox[t]{2mm}{\multirow{16}{*}{\rotatebox[origin=c]{90}{UTKFace}}}

         &ID & 26 & -- & -- & -- & -- & -- & --\\ 
         
         \cdashline{2-9}[1pt/1pt]
         & \multirow{15}{*}{NAS} 
         
          &    25 & 49.7{\scriptsize $\pm$1.8} & 46.8{\scriptsize $\pm$0.9} & 50.0{\scriptsize $\pm$1.2} & 51.5{\scriptsize $\pm$1.0} & 49.5{\scriptsize $\pm$1.5} & 50.1{\scriptsize $\pm$1.7}\\ 
         & &    24 & 52.6{\scriptsize $\pm$1.6} & 51.3{\scriptsize $\pm$0.7} & 50.6{\scriptsize $\pm$2.8} & 52.0{\scriptsize $\pm$1.1} & 52.4{\scriptsize $\pm$1.5} & 53.3{\scriptsize $\pm$1.3}\\ 
         & &    23 & 50.1{\scriptsize $\pm$1.4} & 48.5{\scriptsize $\pm$0.8} & 47.9{\scriptsize $\pm$2.4} & 52.3{\scriptsize $\pm$1.3} & 49.8{\scriptsize $\pm$1.2} & 52.7{\scriptsize $\pm$1.2}\\ 
         & &    22 & 52.5{\scriptsize $\pm$1.6} & 49.7{\scriptsize $\pm$1.0} & 51.2{\scriptsize $\pm$1.7} & 53.2{\scriptsize $\pm$1.7} & 52.3{\scriptsize $\pm$1.1} & 56.5{\scriptsize $\pm$1.3}\\ 
         & &    21 & 53.8{\scriptsize $\pm$1.7} & 51.3{\scriptsize $\pm$1.0} & 49.5{\scriptsize $\pm$2.4} & 52.5{\scriptsize $\pm$1.5} & 53.4{\scriptsize $\pm$1.7} & 55.3{\scriptsize $\pm$1.0}\\ 
         & & 19-20 & 55.9{\scriptsize $\pm$2.1} & 51.8{\scriptsize $\pm$1.0} & 52.2{\scriptsize $\pm$1.5} & 52.4{\scriptsize $\pm$1.6} & 55.7{\scriptsize $\pm$1.8} & 57.9{\scriptsize $\pm$1.2}\\ 
         & & 17-18 & 62.4{\scriptsize $\pm$1.6} & 56.7{\scriptsize $\pm$1.0} & 52.9{\scriptsize $\pm$1.6} & 54.0{\scriptsize $\pm$0.7} & 62.0{\scriptsize $\pm$1.3} & 61.0{\scriptsize $\pm$1.1}\\ 
         & & 15-16 & 69.4{\scriptsize $\pm$1.5} & 60.1{\scriptsize $\pm$2.0} & 59.9{\scriptsize $\pm$2.9} & 56.6{\scriptsize $\pm$1.6} & 68.7{\scriptsize $\pm$1.1} & 70.7{\scriptsize $\pm$0.7}\\ 
         & & 12-14 & 76.1{\scriptsize $\pm$1.3} & 69.2{\scriptsize $\pm$2.3} & 57.7{\scriptsize $\pm$2.6} & 57.7{\scriptsize $\pm$1.6} & 75.5{\scriptsize $\pm$1.1} & 75.3{\scriptsize $\pm$0.9}\\ 
         & &  9-11 & 77.9{\scriptsize $\pm$0.7} & 77.4{\scriptsize $\pm$2.6} & 56.0{\scriptsize $\pm$3.8} & 55.9{\scriptsize $\pm$2.6} & 77.4{\scriptsize $\pm$1.5} & 80.5{\scriptsize $\pm$0.9}\\ 
         & &   7-8 & 77.6{\scriptsize $\pm$0.5} & 78.7{\scriptsize $\pm$2.1} & 53.3{\scriptsize $\pm$4.4} & 56.1{\scriptsize $\pm$1.4} & 77.0{\scriptsize $\pm$1.4} & 82.1{\scriptsize $\pm$1.9}\\
         & &   5-6 & 79.6{\scriptsize $\pm$1.0} & 80.5{\scriptsize $\pm$2.6} & 53.2{\scriptsize $\pm$2.4} & 55.4{\scriptsize $\pm$2.4} & 79.2{\scriptsize $\pm$1.2} & 83.5{\scriptsize $\pm$1.4}\\
         & &   3-4 & 79.8{\scriptsize $\pm$1.2} & 85.5{\scriptsize $\pm$2.4} & 56.4{\scriptsize $\pm$3.2} & 55.4{\scriptsize $\pm$1.2} & 79.0{\scriptsize $\pm$1.5} & 88.6{\scriptsize $\pm$1.0}\\ 
         & &     2 & 79.3{\scriptsize $\pm$0.6} & 88.1{\scriptsize $\pm$1.7} & 53.3{\scriptsize $\pm$2.8} & 53.5{\scriptsize $\pm$0.8} & 78.6{\scriptsize $\pm$1.8} & 88.3{\scriptsize $\pm$1.6}\\ 
         & &     1 & 80.7{\scriptsize $\pm$0.4} & 90.7{\scriptsize $\pm$0.9} & 51.2{\scriptsize $\pm$4.4} & 52.9{\scriptsize $\pm$2.1} & 79.9{\scriptsize $\pm$1.3} & 90.3{\scriptsize $\pm$2.0}\\

        \toprule
        
        
        &\multicolumn{1}{c}{\multirow{2}{*}{Distribution}} & \multicolumn{1}{c}{\multirow{2}{*}{Year}} 
        & \multicolumn{5}{c}{\multirow{1}{*}{CE}}  & \multicolumn{1}{c}{Ours} \\ \cmidrule(lr){4-8}  \cmidrule(lr){9-9} 
        &\multicolumn{1}{c}{} & \multicolumn{1}{c}{} & \multicolumn{1}{c}{MSP} & \multicolumn{1}{c}{ODIN} & \multicolumn{1}{c}{Mahalanobis} & \multicolumn{1}{c}{Gram matrix} &  \multicolumn{1}{c}{Energy score} & \multicolumn{1}{c}{Mahalanobis}

        \\ \midrule
        
        \parbox[t]{2mm}{\multirow{9}{*}{\rotatebox[origin=c]{90}{Amazon Review}}}
        
        & ID & 2005 & -- & -- & -- & -- & -- & --\\
        \cdashline{2-9}[1pt/1pt]
        & \multirow{9}{*}{NAS}
           & 2006 & 49.6{\scriptsize $\pm$0.4} & 49.6{\scriptsize $\pm$0.4} & 51.5{\scriptsize $\pm$0.1} & - & 49.5{\scriptsize $\pm$0.2} & 51.4{\scriptsize $\pm$0.1}\\ 
         
         & & 2007 & 49.0{\scriptsize $\pm$1.4} & 49.0{\scriptsize $\pm$1.4} & 56.8{\scriptsize $\pm$0.2} & - & 47.7{\scriptsize $\pm$0.7} & 56.5{\scriptsize $\pm$0.1}\\ 
         
         & & 2008 & 48.4{\scriptsize $\pm$1.1} & 48.4{\scriptsize $\pm$1.1} & 55.2{\scriptsize $\pm$0.3} & - & 47.5{\scriptsize $\pm$0.5} & 54.9{\scriptsize $\pm$0.1}\\ 
         
         & & 2009 & 48.5{\scriptsize $\pm$1.1} & 48.5{\scriptsize $\pm$1.1} & 53.7{\scriptsize $\pm$0.2} & - & 47.8{\scriptsize $\pm$0.5} & 53.6{\scriptsize $\pm$0.1}\\ 
         
         & & 2010 & 48.1{\scriptsize $\pm$0.2} & 48.7{\scriptsize $\pm$1.2} & 54.0{\scriptsize $\pm$0.5} & - & 47.9{\scriptsize $\pm$0.6} & 53.6{\scriptsize $\pm$0.1}\\ 
         
         & & 2011 & 48.1{\scriptsize $\pm$0.3} & 47.3{\scriptsize $\pm$1.6} & 55.6{\scriptsize $\pm$0.7} & - & 45.7{\scriptsize $\pm$0.4} & 54.9{\scriptsize $\pm$0.0}\\ 
         
         & & 2012 & 45.7{\scriptsize $\pm$0.3} & 45.5{\scriptsize $\pm$2.4} & 63.3{\scriptsize $\pm$0.8} & - & 42.0{\scriptsize $\pm$0.9} & 62.6{\scriptsize $\pm$0.1}\\ 
         
         & & 2013 & 38.2{\scriptsize $\pm$0.6} & 43.0{\scriptsize $\pm$3.5} & 75.5{\scriptsize $\pm$0.5} & - & 38.2{\scriptsize $\pm$0.9} & 75.1{\scriptsize $\pm$0.0}\\ 
         
         & & 2014 & 38.9{\scriptsize $\pm$0.8} & 43.7{\scriptsize $\pm$3.7} & 76.8{\scriptsize $\pm$0.2} & - & 38.8{\scriptsize $\pm$1.0} & 76.7{\scriptsize $\pm$0.1}\\

        \toprule
        
        
        &\multicolumn{1}{c}{\multirow{2}{*}{Distribution}} & \multicolumn{1}{c}{\multirow{2}{*}{Brightness}} 
        & \multicolumn{5}{c}{\multirow{1}{*}{CE}}  & \multicolumn{1}{c}{Ours} \\ \cmidrule(lr){4-8}  \cmidrule(lr){9-9} 
        &\multicolumn{1}{c}{} & \multicolumn{1}{c}{} & \multicolumn{1}{c}{MSP} & \multicolumn{1}{c}{ODIN} & \multicolumn{1}{c}{Mahalanobis} & \multicolumn{1}{c}{Gram matrix} &  \multicolumn{1}{c}{Energy score} & \multicolumn{1}{c}{Mahalanobis}

        \\ \midrule

        \parbox[t]{2mm}{\multirow{16}{*}{\rotatebox[origin=c]{90}{RSNA Bone Age}}}

        &\multirow{5}{*}{NAS} & 0.0 & 93.9{\scriptsize $\pm$4.2} & 100.0{\scriptsize $\pm$0.0} & 99.9{\scriptsize $\pm$0.2} & 99.8{\scriptsize $\pm$0.5} & 92.2{\scriptsize $\pm$4.1} & 100.0{\scriptsize $\pm$0.0}\\ 
        
         & & 0.2 & 71.7{\scriptsize $\pm$3.5} & 89.1{\scriptsize $\pm$6.1} & 93.7{\scriptsize $\pm$1.8} & 68.7{\scriptsize $\pm$2.7} & 71.0{\scriptsize $\pm$4.3} & 98.0{\scriptsize $\pm$1.0}\\ 
         
         & & 0.4 & 55.7{\scriptsize $\pm$2.2} & 64.5{\scriptsize $\pm$4.8} & 79.5{\scriptsize $\pm$4.5} & 54.6{\scriptsize $\pm$3.3} & 55.2{\scriptsize $\pm$2.6} & 89.8{\scriptsize $\pm$2.2}\\ 
         
         & & 0.6 & 52.3{\scriptsize $\pm$1.5} & 53.4{\scriptsize $\pm$3.1} & 64.5{\scriptsize $\pm$5.5} & 51.3{\scriptsize $\pm$1.9} & 52.1{\scriptsize $\pm$1.6} & 74.6{\scriptsize $\pm$2.5}\\  
         
         & & 0.8 & 50.2{\scriptsize $\pm$1.1} & 48.9{\scriptsize $\pm$1.8} & 53.3{\scriptsize $\pm$2.4} & 49.9{\scriptsize $\pm$1.0} & 50.1{\scriptsize $\pm$1.0} & 56.3{\scriptsize $\pm$1.8}\\ 
         \cdashline{2-9}[1pt/1pt]

        & ID & 1.0 & -- & -- & -- & -- & -- & --\\ \cdashline{2-9}[1pt/1pt]

        &\multirow{8}{*}{NAS} 
        
           & 1.2 & 51.6{\scriptsize $\pm$1.2} & 54.9{\scriptsize $\pm$1.7} & 58.5{\scriptsize $\pm$1.9} & 51.4{\scriptsize $\pm$1.0} & 51.4{\scriptsize $\pm$1.3} & 55.0{\scriptsize $\pm$2.4}\\  
        
         & & 1.4 & 54.7{\scriptsize $\pm$2.2} & 63.5{\scriptsize $\pm$3.0} & 69.0{\scriptsize $\pm$3.1} & 54.8{\scriptsize $\pm$1.5} & 54.4{\scriptsize $\pm$2.3} & 65.1{\scriptsize $\pm$3.1}\\ 
         
         & & 1.6 & 58.5{\scriptsize $\pm$3.5} & 71.7{\scriptsize $\pm$4.0} & 76.9{\scriptsize $\pm$4.0} & 59.8{\scriptsize $\pm$1.9} & 57.9{\scriptsize $\pm$3.3} & 75.5{\scriptsize $\pm$3.9}\\ 
         
         & & 1.8 & 61.1{\scriptsize $\pm$3.9} & 80.1{\scriptsize $\pm$4.5} & 82.1{\scriptsize $\pm$5.1} & 62.6{\scriptsize $\pm$2.0} & 60.5{\scriptsize $\pm$3.7} & 82.3{\scriptsize $\pm$4.3}\\ 
         
         & & 2.0 & 62.5{\scriptsize $\pm$4.3} & 87.1{\scriptsize $\pm$4.1} & 85.8{\scriptsize $\pm$5.5} & 64.2{\scriptsize $\pm$3.5} & 61.8{\scriptsize $\pm$4.2} & 87.0{\scriptsize $\pm$4.0}\\  
         
         & & 2.5 & 66.5{\scriptsize $\pm$5.3} & 95.0{\scriptsize $\pm$2.3} & 91.9{\scriptsize $\pm$5.0} & 69.5{\scriptsize $\pm$4.6} & 65.8{\scriptsize $\pm$5.3} & 93.7{\scriptsize $\pm$2.9}\\ 
         
         & & 3.0 & 75.1{\scriptsize $\pm$3.9} & 98.2{\scriptsize $\pm$0.9} & 95.4{\scriptsize $\pm$3.3} & 76.4{\scriptsize $\pm$4.0} & 74.6{\scriptsize $\pm$3.6} & 96.6{\scriptsize $\pm$1.8}\\ 
         
         & & 3.5 & 80.1{\scriptsize $\pm$2.5} & 99.2{\scriptsize $\pm$0.4} & 96.8{\scriptsize $\pm$2.7} & 80.3{\scriptsize $\pm$5.1} & 79.6{\scriptsize $\pm$1.9} & 97.8{\scriptsize $\pm$1.3}\\ 
         
         & & 4.0 & 83.4{\scriptsize $\pm$2.5} & 99.6{\scriptsize $\pm$0.3} & 97.7{\scriptsize $\pm$2.1} & 84.7{\scriptsize $\pm$5.7} & 82.9{\scriptsize $\pm$1.8} & 98.4{\scriptsize $\pm$0.9}\\ 
         
         & & 4.5 & 85.1{\scriptsize $\pm$3.5} & 99.7{\scriptsize $\pm$0.3} & 98.2{\scriptsize $\pm$1.6} & 87.8{\scriptsize $\pm$4.9} & 84.5{\scriptsize $\pm$2.8} & 98.7{\scriptsize $\pm$0.8}\\ 
         \bottomrule
    \end{tabular}}
    \vspace{-2mm}
\caption{NAS detection performance on distributional shifts in three datasets measured by AUROC. 
}
\label{tab:auroc}
\end{table*}


%% file: 060relatedwork.tex
\textbf{Extensive evaluation on distribution shifts. \enskip} 
Recent works have leveraged the distribution shifts for OOD detection and uncertainty estimation.
\citet{rabanser2019failing} primarily focus on detecting the entire shifted distribution, not on detecting a single shifted sample. 
\citet{ovadia2019can} quantify the predictive uncertainty and investigate the quality of the calibration under the dataset shift.
\citet{practical20} aims to predict the classification accuracy of a shifted distribution by utilizing the average OOD score of the distribution.
\citet{godin} also consider the distribution shift with preserved label space.
However, \citet{godin} focus on the shifted distribution from different domain ($\textit{e.g.}$, real images vs. sketches ).
To the best of our knowledge, none of these works conduct an \textit{in-depth analysis} on the performance of existing OOD detection methods on samples shifted based on natural attributes.

\textbf{OOD detection methods. \enskip}
Post-hoc OOD detection methods~\citep{hendrycks2017base,liang2017enhancing,mahala18,sastry2021gram} that utilize the classification models to obtain the OOD scores have achieved remarkable performance.
All of these works aim to detect shifted sample which might affect decision making system in terms of confidence and hidden representation.
Other OOD detection approach is to train the generative model~\citep{ren2019likelihood,choi2018waic, mahmood2020multiscale} on the training distribution and estimate the density of OOD samples in test time. 
While these approaches are viable, it is not directly related with the downstream task but aims to detect features which is different from training distribution.
In this paper, we mainly focus on methods that utilize the classification models to detect OOD samples since we are mainly interested in samples that affect decision-making systems.

\textbf{Model uncertainty. \enskip} A number of previous works measure model uncertainty using various methods such as Bayesian neural networks~\citep{blundell2015weight}, Monte Carlo dropout~\citep{gal2016dropout}, and deep ensembles~\citep{lakshminarayanan2017simple}.
Note that technically, model uncertainty can be used to detect NAS samples, especially for NAS categories 1 and 3, since sampling model weights from the function space can be seen as redrawing the decision boundary, and NAS samples in categories 1 and 3 will be affected heavily by this process.
Model uncertainty, however, aims to capture the uncertainty in the model weights rather than detecting OOD samples, making the two rather independent research directions.

%% file: 070conclusions.tex
To enhance the reliability of decision-making systems, we define a new task, Natural Attribute-based Shift (NAS) detection, that aims to detect the samples shifted by a natural attribute.
We introduce NAS detection benchmark datasets by adjusting the natural attributes present in the existing datasets.
Through extensive evaluation of existing OOD detection methods on NAS datasets, we observe inconsistent performance depending on the nature of NAS samples.
Then, we analyze the inconsistency by probing the relationship between the location of NAS samples and the performance of existing OOD detection methods.
Based on this observation, we suggest a simple remedy to help Mahalanobis OOD detection method to have consistent performance across all NAS categories.
We hope our dataset and task inspire fellow researchers to investigate practical methods for identifying NAS, which is crucial for deploying the prediction models in real-world systems.

%% file: main_appendix.tex
\appendix
\addcontentsline{toc}{section}{Appendix}

\renewcommand \thepart{}
\renewcommand \partname{}
\doparttoc
\faketableofcontents 

\part{} 
\parttoc

\section{Details of OOD detection methods}
\label{sec:supp_ood_details}
\input{900details_of_method}

\section{Analysis on the Text Dataset}
\label{sec:supp_dataset_details}
\input{901details_of_dataset}

\section{Implementation Details}
\label{sec:supp_implementation_details}
\input{902implemetation_details}

\section{Ablation Studies on Distance and Entropy Loss}
\label{sec:supp_ablations}
\input{903ablation}

\newpage
\clearpage
\section{Comparison with Additional Baselines}
\label{sec:supp_additional_baselines}
\input{904comparison_baselines}

\section{Quantitative Results with Other Metrics}
\label{sec:supp_other_metrics}

\input{905metrics}

%% file: 900details_of_method.tex
In this section, we describe the three  post-hoc and task-agnostic OOD detection methods in detail, focusing mainly on their formulation and how each method assigns an OOD score to an input sample.

\subsection{Maximum of Softmax Probability (MSP)}
In this method, the maximum of softmax probability is considered as confidence score~\citep{hendrycks2017base}.
Formally, we calculate the maximum softmax probability as follows:

\begin{equation*}
\centering
S_\text{MSP}(\textbf{x}) = \max_c \frac{\exp{\left(\textbf{z}^{(c)}\right)}}{\sum_{j=1}^C \exp{\left(\textbf{z}^{(j)}\right)}},
\end{equation*}
where $C$ is the number of target classes, $c$ is the index of a class, and $\textbf{z}^{(j)}$ denotes $j^{th}$ attribute of the feature in the logit layer.

\subsection{ODIN}

ODIN \citep{liang2017enhancing} utilized two well established techniques, namely temperature scaling and input preprocessing to increase the difference between softmax scores of in-distribution and OOD samples.
Temperature scaling was originally proposed in \citet{Hinton2015DistillingTK} to distill the knowledge in neural networks and was later adopted widely in classification tasks to calibrate confidence of prediction\citep{Guo2017OnCO}. In addition to temperature scaling, the input is preprocessed in order to increase the softmax score of given input by adding small perturbations which are obtained by back-propagating the gradient of the loss with respect to the input.  More specifically, ODIN is computed as follows:

\begin{equation*}
\centering
S(\textbf{x};T) = \max_c \frac{\exp{\left(\textbf{z}^{(c)}/T\right)}}{\sum_{j=1}^C \exp{\left(\textbf{z}^{(j)}/T\right)}},
\end{equation*}
where $T \in \mathbb{R}^{+}$ is the temperature scaling parameter, $C$ is the number of target classes, $c$ is the index of class, and $\textbf{z}^{(j)}$ denotes $j^{th}$ attribute of the logit layer features of input $\textbf{x}$. During training, $T$ is set to $1$.

For OOD detection, the input is first pre-processed as follows:
\begin{equation*}
\centering 
\tilde{\textbf{x}}=\textbf{x}-\varepsilon \operatorname{sign}\left(-\nabla_{\boldsymbol{x}} \log S(\textbf{x} ; T)\right), 
\end{equation*}
where $\varepsilon$ represents the magnitude of perturbation.

Next, the network calculates the calibrated softmax score of the preprocessed input as follows:
\begin{equation*}
\centering
S_\text{ODIN}(\textbf{x};T) = \max_c \frac{\exp{\left(\tilde{\textbf{z}}^{(c)}/T\right)}}{\sum_{j=1}^C \exp{\left(\tilde{\textbf{z}}^{(j)}/T\right)}},
\end{equation*}
where $\tilde{\textbf{z}}^{(j)}$ denotes $j^{th}$ attribute of the logit layer features of the preprocessed input $\tilde{\textbf{x}}$ . 

Lastly, the modified softmax score is compared to a threshold value $\delta$. If the score is greater than the threshold, then the input is classified as ID sample and otherwise OOD. 
Originally, $T$, $\varepsilon$, and $\delta$ are hyperparameters and are selected such that the false positive rate (FPR) at true positive rate (TPR) 95\% is minimized on the validation OOD dataset. However, the performance saturates when $T$ is greater than $1000$ and therefore, in general, a large value of $T$ is preferred. Following this, in this paper, we fix $T=1000$ for our experiments.

\subsection{Mahalanobis Detector}
To obtain Mahalanobis distance-based OOD score~\citep{mahala18} of a sample, we calculate the mahalanobis distance from the clusters of classes to the sample. 
Then, the distance from the closest class is chosen as the confidence score. 

Specifically, the Mahalanobis score of an input $\textbf{x}$ is defined as
\begin{equation*}
\centering
    S_{\text{mahala}}(\textbf{x}) = \mathbf\sum_{l}{\alpha_l \max_{c} -\left(f^{l}_{\theta}\left(\textbf{x}\right) - \mu_{c,l}\right)^\top \mathbf{\Sigma}^{-1}_l \left(f^{l}_{\theta}\left(\textbf{x}\right) - \mu_{c,l}\right) },
\end{equation*}
where $c$ and $l$ are the class and layer index, respectively, $f_\theta^l$ is the $l^{th}$ layer's feature representation of an input $\textbf{x}$, $\mu_{c,l}$ and  $\mathbf{\Sigma}_l$ are their class mean vector and tied covariance of the training data, correspondingly. 

Note that \textbf{ODIN} and \textbf{Mahalanobis Detector} assume the availability of OOD validation dataset.
However, some recent works~\citep{shafei18,hfree} report that this assumption limits the OOD detection generalizability since a model is biased towards an OOD validation set.
In response, this paper validate the performance of OOD methods in the version that does not require to tune with OOD validation dataset. We perform the performance of ODIN as \cite{godin}. 
We do not perform Mahalanobis Detector ensembling over on all layers with the optimal linear combination which requires explicit OOD data. Instead, we perform two version that use only the penultimate layer of hidden representation and sum uniformly over all layers. 
Therefore, for all our experiments, we use modified OOD detection methods that do not require the OOD validation dataset.

%% file: 901details_of_dataset.tex
\begin{figure}[h!]
\centering
    \includegraphics[width=\textwidth]{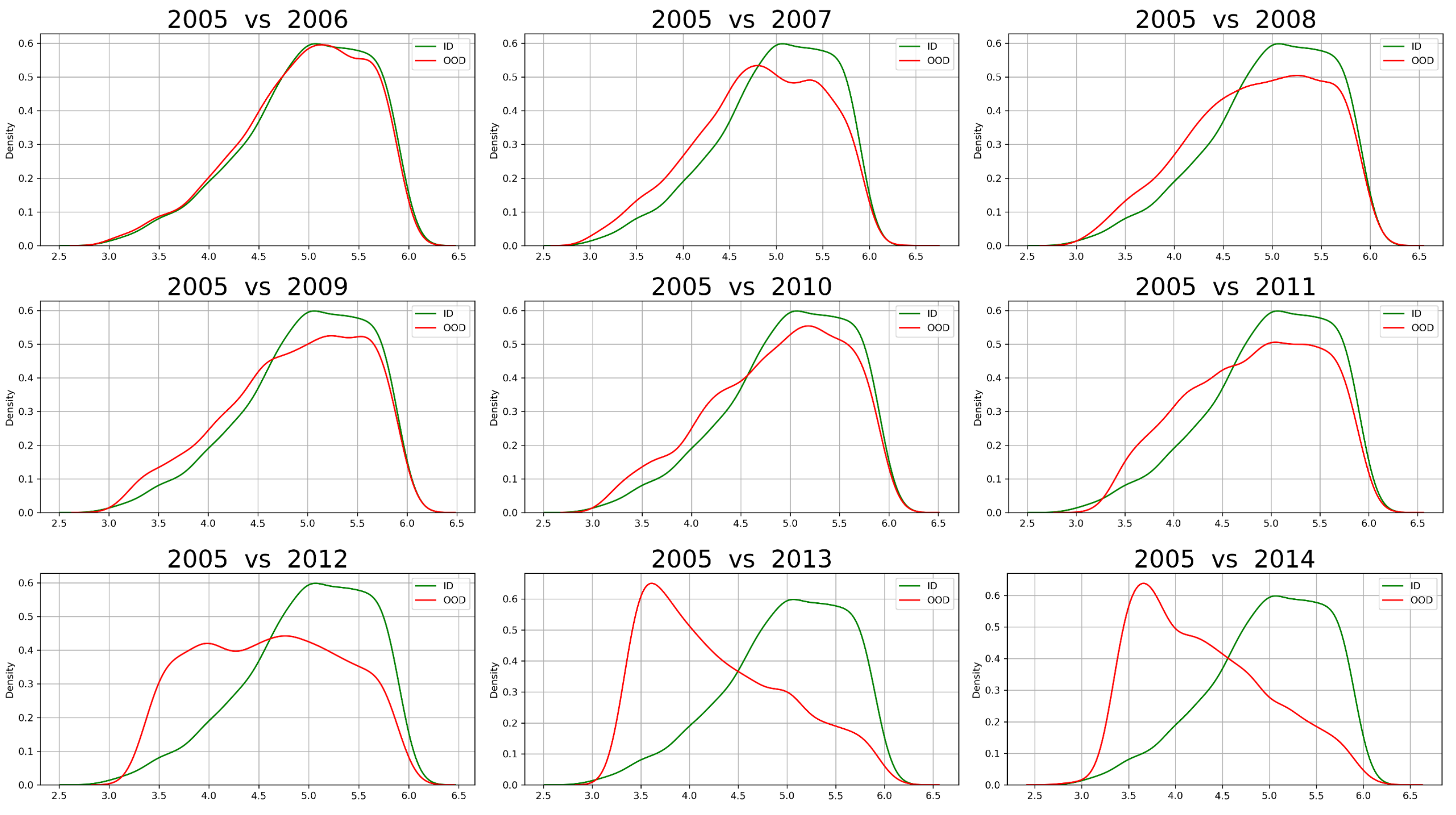}
    \caption{The density plot of sequence length per year. The x-axis represents the length with log scaling.}
    \label{fig:hist_length}
\vspace{-2mm}
\end{figure}
\begin{figure}[h!]
\centering
    \includegraphics[width=\textwidth]{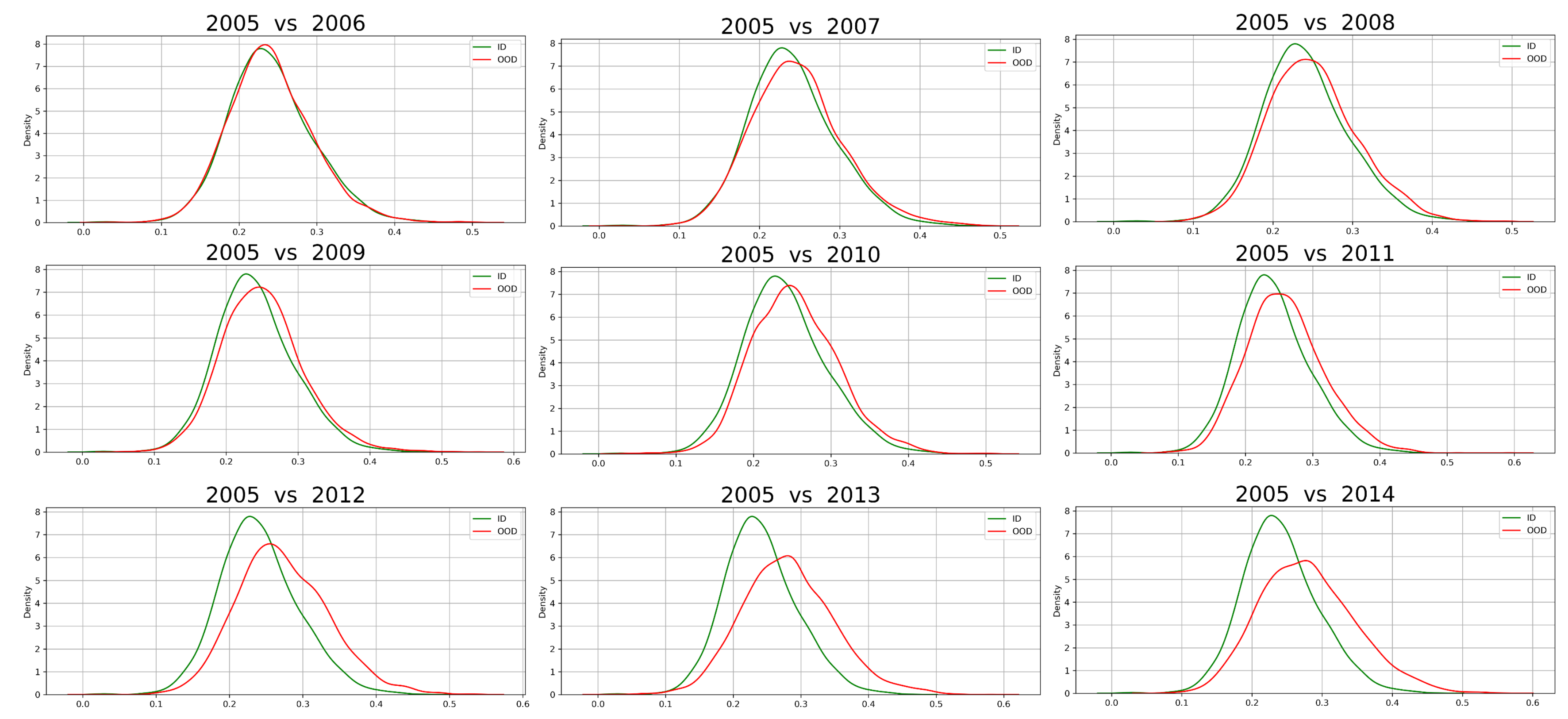}
    \caption{The density plot of the ratio of important words per sequence.}
    \label{fig:hist_ratio}
\vspace{-4mm}
\end{figure}

In this section, we provide a detailed analysis of the text dataset.
We use the Amazon Review~\citep{HeM16, McAuleyTSH15} dataset.
We consider the product category "book" and conducted an analysis to see the impact of time on product reviews. 
We then performed an analysis to see the impact of time on the length of the reviews. 
Figure~\ref{fig:hist_length} presents a comparison between density plot of review length for each year from $2006$ to $2014$. 
We observe that as we move ahead in time,  the length of the reviews gradually reduces. 
Further, Figure~\ref{fig:hist_ratio} presents the distribution of the average ratio of important words in a sequence by year. 
First, we train a model to classify sentiment polarity using Catboost~\citep{prokhorenkova2017catboost} using document term frequency vector. 
We then extract top-100 important words using feature importance. Lastly, we manually select important words over 100 words. 
We find that the distribution of ratio of important words in a sequence gradually increase over time. Based on this analysis, we figure out that the feature related with downtream task is shifted by time.

%% file: 902implemetation_details.tex
In this section, we describe the training details. Followed by it, we describe the algorithm we used to select the hyperparameters in our suggested modification to the loss function. Then, we provide the links to download the datasets.

\subsection{Training Details and Computing Infrastructure}

\textbf{Image.}
We used ResNet18~\citep{he2016deep} that is pretrained with ImageNet~\citep{5206848} to train a gender classifier using a UTKFace NAS dataset. After the average pooling layer, we trained a fully-connected network composed of $2$ hidden layers which have $128$ and $2$ units respectively with a relu activation. The network is trained for 100 epochs with a batch size of 64. We used stochastic gradient descent with Adam optimizer~\citep{kingma2014adam} and set  learning rate as $3e-5$. For data augmentation, the technique of SimCLR~\citep{chen2020simple} is used for all the experiments. 

\textbf{Text.}
We perform experiment with Transformer network with $4$ layers. The network is trained for $10$ epochs with batch size of $128$. 
We used Adam as an optimizer and set the learning rate as $3e-6$.

\textbf{Medical.}
 We use a ResNet18~\citep{he2016deep} model, pretrained on ImageNet~\citep{5206848}, and add two fully-connected layers containing $128$ and $2$ hidden units with a relu activation. We train the network to classify gender given the x-ray image.
Each model is trained for $30$ epochs using SGD optimizer with a learning rate of $0.01$ and momentum of $0.9$, using a batch size of $64$. 

All the experiments are conducted on a single GeForce RTX $3090$ GPU with $24$GB memory. 
The results are measured by computing mean and standard deviation across $5$ trials upon randomly chosen seeds.

\subsection{Hyperparameter Tuning}
Our suggested modification to the training loss mainly comprises of \textit{distance loss} ($\mathcal{L}_\text{dist}$)  and a \textit{entropy loss} ($\mathcal{L}_\text{entropy}$). 
More formally, the training loss is given by: 
\begin{equation}
    \mathcal{L}_\text{total} = \mathcal{L}_\text{CE} + \mathcal{L}_\text{dist} + \mathcal{L}_\text{entropy}.
    \label{eq:supp_total_loss}
\end{equation}
The \textit{entropy loss} comprises of two terms and is defined as:
\begin{equation}
    \mathcal{L}_\text{entropy} = \lambda_{2}\frac{D}{\sum_{i} \mathrm{Var}(\mathbf{z}_i)} + \lambda_{3}\frac{1}{\binom{D}{2}}\sum_{i}\sum_{j \neq i} C_{ij}^2,
    \label{eq:supp_entropy}
\end{equation}
where $\lambda_2>0$ and $\lambda_3>0$ are hyperparameters, $\mathbf{z} \in \mathbb{R}^{D}$ is the feature representation in the latent feature space, and $C_{ij}$ is the correlation coefficient between $i$th and $j$th dimension of the feature space.
For more details, please refer to Section~\ref{sec:new_method}.

For simplicity, in this section, we will refer the first term of \textit{entropy loss} as \textit{variance loss}, and the second term as \textit{correlation loss}.
To obtain the hyperparameters of different terms in our loss function, we explore the value of $\lambda_2$ in $\left[0.01, 0.1, 1.0, 10.\right]$ and $\lambda_3$ in $\left[0.0001, 0.001, 0.01, 0.1, 1.0\right]$.
The hyperparameter corresponding the \textit{distance loss}, $\lambda_{1}$, is set as 0.1, and then the hyperparameters of the \textit{variance loss} and the \textit{correlation loss} are chosen by a simple algorithm.

\textbf{Algorithm:}
We now describe the algorithm we used to find the hyperparameters of \textit{variance} and \textit{correlation loss}.
First, we calculate the harmonic mean of the \textit{variance loss} and \textit{correlation loss} using our training dataset. Next, we select the hyperparameters  with the lowest value of harmonic mean.
Then, to ensure that \textit{entropy loss} prevents the feature space collapsing problem, we apply the singular value decomposition (SVD) in the penultimate features and test if the sum of singular values except the two largest values is improved by the \textit{entropy loss} or not. Concretely, to prove the enhancement, we compare the values with those of the model trained with the $\mathcal{L}_{\text{CE}} + \lambda_1\mathcal{L}_{\text{dist}}$.
If the selected hyperparameters do not improve the values, we reject them and investigate the hyperparameters that have the next lowest harmonic mean. 
The hyperparameters satisfying the above steps are selected as the hyperparameters of our loss function.

Note that we reject the hyperparameters if they significantly degrade the classification accuracy.
In image domain, we applied the algorithm and obtained $0.1$, $0.1$, $0.0001$ as hyperparameters of
$\lambda_1$, $\lambda_2$, and $\lambda_3$, respectively.
In the medical domain, we obtain the hyperparameter corresponding to $\lambda_3$ as $1.0$, and the other hyperparameters are the same as those of the image domain.
In the text domain, $10.$ and $1.0$ are used for the $\lambda_2$ and $\lambda_3$. The \textit{distance loss} is set by $0.1$.  
 
\subsection{Links for Datasets}
In our work, we use the openly available datasets, namely, UTKFace dataset\footnote{\url{ https://susanqq.github.io/UTKFace/}}, Amazon Review dataset\footnote{\url{https://jmcauley.ucsd.edu/data/amazon/}}, and RSNA Bone Age dataset\footnote{\url{https://www.rsna.org/education/ai-resources-and-training/ai-image-challenge/rsna-pediatric-bone-age-challenge-2017}}.

%% file: 903ablation.tex
\input{tabels/supp/ablation1}

In this section, we conduct an ablation study to investigate the effect of each proposed term in the loss function in ~\eqref{eq:supp_total_loss}. 
Table~\ref{tab:ablation} presents an ablation study to find the effect of the \textit{entropy loss} and \textit{distance loss} in the proposed modification to training loss.
In UTKFace, when training with solely attaching $\mathcal{L}_\text{dist}$ or $\mathcal{L}_\text{entropy}$ to $\mathcal{L}_\text{CE}$, the NAS detection performance using mahalanobis OOD detection method improves compared to training with only $\mathcal{L}_\text{CE}$, but the performance does not monotonically increase in both cases along with the variation of Age.
However, when both $\mathcal{L}_\text{dist}$ and $\mathcal{L}_\text{entropy}$ are used, there is a large improvement in performance with the results being monotonically increasing as we move from ID towards NAS, which implies that these two losses are mutually beneficial.
In the text and medical domains, although there is a little degradation of AUROC when only $\mathcal{L}_\text{dist}$ is added, the performance is improved by using $\mathcal{L}_\text{entropy}$ in the training objective.

Further, to analyze the impact of each loss term to the feature-level representations, we perform a singular value decomposition (SVD) in the penultimate layer of ResNet18 trained on the UTKFace dataset, similar to \citet{verma2019manifold}.
When trained with $\mathcal{L}_\text{CE}$ only, the sum of the two largest singular values was $617.47$, while the sum of the remaining singular values was $785.54$.
When trained with $\mathcal{L}_\text{CE} + \mathcal{L}_\text{dist}$, the sum of the two largest singular values increased ($4481.88$), while the sum of the remaining singular values decreased ($371.86$).

As discussed in Section~\ref{sec:new_method} in the main paper, this indicates that the addition of the \textit{distance loss} is likely to collapse the latent feature space (\textit{i.e.} rank-deficient), thus possibly decreasing the model's sensitivity to NAS samples. 
For example, if a model is trained to classify apples and bananas based only on the color, it will not be able to tell that a firetruck is a NAS sample.
When we train the classifier with $\mathcal{L}_\text{entropy}$ added, the problem is effectively alleviated.
The sum of the two largest singular value is reduced from $4481.88$ to $2960.63$, and the sum of the remaining singular values increased from $371.86$ to $513.13$.

%% file: tabels/supp/ablation1.tex
\begin{table*}[h]
\small
\centering
\resizebox{\linewidth}{!}{
    \begin{tabular}{c c c  c  c  c c  }
        \toprule
        & Distribution & Age & $\mathcal{L}_\text{CE}$ & $\mathcal{L}_\text{CE}$  + $\mathcal{L}_\text{dist}$ & $\mathcal{L}_\text{CE}$ + $\mathcal{L}_\text{entropy}$ &  $\mathcal{L}_\text{CE}$  + $\mathcal{L}_\text{dist}$ + $\mathcal{L}_\text{entropy}$
        
        \\ \midrule
        
        \parbox[t]{2mm}{\multirow{16}{*}{\rotatebox[origin=c]{90}{UTKFace}}}
      
         & ID & 26 & -- & -- & -- & --\\ 
         
         \cdashline{2-7}[1pt/1pt]
         & \multirow{15}{*}{NAS} 
          &    25 & 50.0{\scriptsize $\pm$1.2} & 50.3{\scriptsize $\pm$1.0} & 56.8{\scriptsize $\pm$2.4} & 50.1{\scriptsize $\pm$1.7}\\ 
         & &    24 & 50.6{\scriptsize $\pm$2.8} & 50.8{\scriptsize $\pm$0.8} & 54.6{\scriptsize $\pm$1.7} & 53.3{\scriptsize $\pm$1.3}\\ 
         & &    23 & 47.9{\scriptsize $\pm$2.4} & 48.8{\scriptsize $\pm$1.2} & 58.1{\scriptsize $\pm$3.3} & 52.7{\scriptsize $\pm$1.2}\\ 
         & &    22 & 51.2{\scriptsize $\pm$1.7} & 53.4{\scriptsize $\pm$1.0} & 61.5{\scriptsize $\pm$2.8} & 56.5{\scriptsize $\pm$1.3}\\ 
         & &    21 & 49.5{\scriptsize $\pm$2.4} & 51.7{\scriptsize $\pm$2.3} & 56.0{\scriptsize $\pm$3.6} & 55.3{\scriptsize $\pm$1.0}\\ 
         & &    19-20 & 52.2{\scriptsize $\pm$1.5} & 54.8{\scriptsize $\pm$2.0} & 56.6{\scriptsize $\pm$3.1} & 57.9{\scriptsize $\pm$1.2}\\ 
         & &    17-18 & 52.9{\scriptsize $\pm$1.6} & 58.5{\scriptsize $\pm$2.4} & 62.0{\scriptsize $\pm$4.5} & 61.0{\scriptsize $\pm$1.1}\\ 
         & &    15-16 & 59.9{\scriptsize $\pm$2.9} & 70.0{\scriptsize $\pm$3.4} & 66.6{\scriptsize $\pm$4.2} & 70.7{\scriptsize $\pm$0.7}\\ 
         & &    12-14 & 57.7{\scriptsize $\pm$2.6} & 67.4{\scriptsize $\pm$5.5} & 65.2{\scriptsize $\pm$5.6} & 75.3{\scriptsize $\pm$0.9}\\ 
         & &    9-11 & 56.0{\scriptsize $\pm$3.8} & 66.5{\scriptsize $\pm$7.6} & 67.7{\scriptsize $\pm$5.8} & 80.5{\scriptsize $\pm$0.9}\\ 
         & &    7-8 & 53.3{\scriptsize $\pm$4.4} & 64.1{\scriptsize $\pm$7.9} & 69.3{\scriptsize $\pm$7.3} & 82.1{\scriptsize $\pm$1.9}\\ 
         & &    5-6 & 53.2{\scriptsize $\pm$2.4} & 63.0{\scriptsize $\pm$8.0} & 64.3{\scriptsize $\pm$5.1} & 83.5{\scriptsize $\pm$1.4}\\ 
         & &    3-4 & 56.4{\scriptsize $\pm$3.2} & 64.3{\scriptsize $\pm$10.1} & 65.8{\scriptsize $\pm$5.1} & 88.6{\scriptsize $\pm$1.0}\\ 
         & &    2 & 53.3{\scriptsize $\pm$2.8} & 58.6{\scriptsize $\pm$10.1} & 57.7{\scriptsize $\pm$4.2} & 88.3{\scriptsize $\pm$1.6}\\ 
         & &    1 & 51.2{\scriptsize $\pm$4.4} & 58.1{\scriptsize $\pm$10.2} & 55.6{\scriptsize $\pm$5.7} & 90.3{\scriptsize $\pm$2.0}\\

        \toprule
        & Distribution & Year & $\mathcal{L}_\text{CE}$ & $\mathcal{L}_\text{CE}$  + $\mathcal{L}_\text{dist}$ & $\mathcal{L}_\text{CE}$ + $\mathcal{L}_\text{entropy}$ &  $\mathcal{L}_\text{CE}$  + $\mathcal{L}_\text{dist}$ + $\mathcal{L}_\text{entropy}$

        \\ \midrule
        
        \parbox[t]{2mm}{\multirow{9}{*}{\rotatebox[origin=c]{90}{Amazon Review}}}
        
       & ID & 2005 & -- & -- & -- & --\\ 
        \cdashline{2-7}[1pt/1pt]
        & \multirow{9}{*}{NAS}

          & 2006 & 51.5{\scriptsize $\pm$0.1} & 51.5{\scriptsize $\pm$0.1} & 51.4{\scriptsize $\pm$0.1} & 51.4{\scriptsize $\pm$0.1}\\ 
        & & 2007 & 56.8{\scriptsize $\pm$0.2} & 56.3{\scriptsize $\pm$0.1} & 56.6{\scriptsize $\pm$0.1} & 56.5{\scriptsize $\pm$0.1}\\ 
        & & 2008 & 55.2{\scriptsize $\pm$0.3} & 54.8{\scriptsize $\pm$0.1} & 54.9{\scriptsize $\pm$0.1} & 54.9{\scriptsize $\pm$0.1}\\ 
        & & 2009 & 53.7{\scriptsize $\pm$0.2} & 53.6{\scriptsize $\pm$0.1} & 53.6{\scriptsize $\pm$0.1} & 53.6{\scriptsize $\pm$0.1}\\ 
        & & 2010 & 54.0{\scriptsize $\pm$0.5} & 53.6{\scriptsize $\pm$0.1} & 53.6{\scriptsize $\pm$0.1} & 53.6{\scriptsize $\pm$0.1}\\ 
        & & 2011 & 55.6{\scriptsize $\pm$0.7} & 54.8{\scriptsize $\pm$0.1} & 54.9{\scriptsize $\pm$0.0} & 54.9{\scriptsize $\pm$0.0}\\ 
        & & 2012 & 63.3{\scriptsize $\pm$0.8} & 62.5{\scriptsize $\pm$0.1} & 62.6{\scriptsize $\pm$0.1} & 62.6{\scriptsize $\pm$0.1}\\ 
        & & 2013 & 75.5{\scriptsize $\pm$0.5} & 74.7{\scriptsize $\pm$0.1} & 75.1{\scriptsize $\pm$0.1} & 75.1{\scriptsize $\pm$0.0}\\ 
        & & 2014 & 76.8{\scriptsize $\pm$0.2} & 76.3{\scriptsize $\pm$0.1} & 76.8{\scriptsize $\pm$0.1} & 76.7{\scriptsize $\pm$0.1}\\

        \toprule
        & Distribution & Brightness & $\mathcal{L}_\text{CE}$ & $\mathcal{L}_\text{CE}$  + $\mathcal{L}_\text{dist}$ & $\mathcal{L}_\text{CE}$ + $\mathcal{L}_\text{entropy}$ &  $\mathcal{L}_\text{CE}$  + $\mathcal{L}_\text{dist}$ + $\mathcal{L}_\text{entropy}$

        \\ \midrule
        
        \parbox[t]{2mm}{\multirow{16}{*}{\rotatebox[origin=c]{90}{RSNA Bone Age}}}

        &\multirow{5}{*}{NAS} & 0.0 & 99.9{\scriptsize $\pm$0.2} & 99.9{\scriptsize $\pm$0.2} & 100.0{\scriptsize $\pm$0.0} & 100.0{\scriptsize $\pm$0.0}\\ 
        
         & & 0.2 & 93.7{\scriptsize $\pm$1.8} & 94.0{\scriptsize $\pm$2.3} & 96.9{\scriptsize $\pm$1.2} & 98.0{\scriptsize $\pm$1.0}\\ 
         
         & & 0.4 & 79.5{\scriptsize $\pm$4.5} & 74.4{\scriptsize $\pm$6.3} & 88.4{\scriptsize $\pm$3.6} & 89.8{\scriptsize $\pm$2.2}\\ 
         
         & & 0.6 & 64.5{\scriptsize $\pm$5.5} & 60.9{\scriptsize $\pm$4.1} & 74.5{\scriptsize $\pm$3.7} & 74.6{\scriptsize $\pm$2.5}\\ 
         
         & & 0.8 & 53.3{\scriptsize $\pm$2.4} & 51.9{\scriptsize $\pm$1.6} & 56.7{\scriptsize $\pm$2.3} & 56.3{\scriptsize $\pm$1.8}\\ 
         \cdashline{2-7}[1pt/1pt]

       & ID & 1.0 & -- & -- & -- & --\\ 
        \cdashline{2-7}[1pt/1pt]
        
        &\multirow{8}{*}{NAS} 
        
           & 1.2 & 58.5{\scriptsize $\pm$1.9} & 57.8{\scriptsize $\pm$2.8} & 57.4{\scriptsize $\pm$1.8} & 55.0{\scriptsize $\pm$2.3}\\ 
        
         & & 1.4 & 69.0{\scriptsize $\pm$3.1} & 67.1{\scriptsize $\pm$4.4} & 68.6{\scriptsize $\pm$1.7} & 65.1{\scriptsize $\pm$3.1}\\ 
         
         & & 1.6 & 76.9{\scriptsize $\pm$4.0} & 75.4{\scriptsize $\pm$5.0} & 78.5{\scriptsize $\pm$1.9} & 75.5{\scriptsize $\pm$3.9}\\ 
         
         & & 1.8 & 82.1{\scriptsize $\pm$5.1} & 81.4{\scriptsize $\pm$4.8} & 84.3{\scriptsize $\pm$1.9} & 82.3{\scriptsize $\pm$4.3}\\ 
         
         & & 2.0 & 85.8{\scriptsize $\pm$5.5} & 86.3{\scriptsize $\pm$3.8} & 88.2{\scriptsize $\pm$1.6} & 87.0{\scriptsize $\pm$4.0}\\ 
         
         & & 2.5 & 91.9{\scriptsize $\pm$5.0} & 93.3{\scriptsize $\pm$3.1} & 94.0{\scriptsize $\pm$1.3} & 93.7{\scriptsize $\pm$2.9}\\ 
         
         & & 3.0 & 95.4{\scriptsize $\pm$3.3} & 96.8{\scriptsize $\pm$1.6} & 96.3{\scriptsize $\pm$1.0} & 96.6{\scriptsize $\pm$1.8}\\ 
         
         & & 3.5 & 96.8{\scriptsize $\pm$2.7} & 98.4{\scriptsize $\pm$1.0} & 97.4{\scriptsize $\pm$0.8} & 97.8{\scriptsize $\pm$1.3}\\
         
         & & 4.0 & 97.7{\scriptsize $\pm$2.1} & 99.0{\scriptsize $\pm$0.7} & 97.9{\scriptsize $\pm$0.8} & 98.4{\scriptsize $\pm$0.9}\\ 
         
         & & 4.5 & 98.2{\scriptsize $\pm$1.6} & 99.2{\scriptsize $\pm$0.6} & 98.2{\scriptsize $\pm$0.8} & 98.7{\scriptsize $\pm$0.8}  \\
         \bottomrule
         
    \end{tabular}
}
\caption{NAS detection performance on three NAS datasets measured with Mahalanobis OOD detection method using AUROC. 
}
\label{tab:ablation}
\vspace{-3mm}
\end{table*}

%% file: 904comparison_baselines.tex
\input{tabels/supp/recentbaselines_acc}

\input{tabels/supp/recentbaselines}

In this section, we investigate NAS detection performance using additional baselines, including recent work, and compare them with those of the model trained with our modified loss on three NAS datasets.
Firstly, as we mentioned in the main paper, we evaluate the model trained with cross-entropy loss with ensembled version of Mahalanobis detector~\citep{mahala18} and the method that utilizes the gram matrix in OOD detection~\citep{sastry2021gram}.

We also consider the other recent baselines for the comparison.
Recently, \citet{sehwag2021ssd} propose a unified framework to leverage self-supervised learning for OOD detection. We set the method as our baseline and denote this baseline as SSD$^{+}$.
The experiment on SSD$^{+}$ requires a data augmentation strategy. In the Image domain, we augment the training dataset as~\citet{chen2020simple}. We construct an augmented training dataset using a back-translation model for text experiments as~\citep{fang2020cert,gunel2020supervised}. We train SSD$^{+}$ for 100 epochs for UTKFace and RSNA Bone age dataset and 50 epochs for Amazon Review dataset.

Some recent methods~\citep{godin, hfree} exploit the cosine similarity to detect OOD samples.
\citet{godin} suggest to decompose confidence scores and to modify input preprocessing method in ODIN~\citep{liang2017enhancing}. For OOD detection, they calculate the cosine similarity between the features from the penultimate layer and the class specific weights and use the maximum value as an OOD score.
\citet{hfree} also propose to compute cosine similarity between the class weight vector and the penultimate layer feature of each input to obtain OOD scores. 
Specifically, they train the model with cross entropy loss as standard classification model, but logit value is set by the scaled cosine similarity value.
After training, the cosine similarity values between the penultimate layer feature of the input and the weight vector for each class are calculated.
The maximum cosine similarity value is used as the OOD score for the samples.
We name the methods as GODIN and Scaled Cosine in Table~\ref{tab:recent_baseline}, respectively.

NAS detection performance of our method and other baselines are provided in Table~\ref{tab:recent_baseline}.
In UTKFace dataset, GODIN, Scaled Cosine, and ensembled version of Mahalanobis detector have relatively low detection performance than the other baselines.
Although Gram matrix shows reasonable detection performance in UTKFace and RSNA Bone Age NAS groups, this method is hard to applied to the Amazon Book Review dataset because the notion of gram matrix is vague in the text domain.
In three NAS categories, SSD+ has the most robust detection performance among the five baselines, however, our simple remedy in training loss help Mahalanobis detector to demonstrate improved performance particularly under the UTKFace NAS dataset.
In conclusion, when compared with the five more recent baselines, our method shows robust and high detection performance in all the three NAS datasets, regardless of the domains. 

%% file: tabels/supp/recentbaselines_acc.tex
\begin{table*}[h]
\footnotesize
\centering
\resizebox{\linewidth}{!}{
    \begin{tabular}{c c c c  c  c c }
        \toprule

        Dataset & CE & GODIN & Scaled Cosine & SSD+ & Ours
        
        \\ \midrule
        UKTFace & 94.6{\scriptsize $\pm$06} & 94.7{\scriptsize $\pm$0.8} & 94.4{\scriptsize $\pm$0.7} & 94.2{\scriptsize $\pm$0.5} & 94.1{\scriptsize $\pm$0.7}\\

         Amazon Review& 85.3{\scriptsize $\pm$0.9} & 85.7{\scriptsize $\pm$0.7} & 85.4{\scriptsize $\pm$0.8} & 83.8{\scriptsize $\pm$0.6} & 84.0{\scriptsize $\pm$0.9} \\
         
         RSNA Bone Age&  93.0{\scriptsize $\pm$0.6} & 92.0{\scriptsize $\pm$1.4} & 93.1{\scriptsize $\pm$1.0} & 71.9{\scriptsize $\pm$6.6} & 92.3{\scriptsize $\pm$0.3} \\

        \bottomrule
    \end{tabular}}
\caption{Classification accuracy of the in-distribution in three NAS datasets.
}
\label{tab:recent_baseline_acc}
\vspace{-3mm}
\end{table*}

%% file: tabels/supp/recentbaselines.tex
\begin{table*}[h]
\small
\centering
\resizebox{\linewidth}{!}{
    \begin{tabular}{c c  c  c  c c  c c c c }
        \toprule
        
        &\multicolumn{1}{c}{\multirow{2}{*}{Distribution}} & \multicolumn{1}{c}{\multirow{2}{*}{Age}} 
        & \multicolumn{5}{c}{\multirow{1}{*}{Baselines}} & \multicolumn{2}{c}{Ours} \\ \cmidrule(lr){4-8}  \cmidrule(lr){9-10} 
        &\multicolumn{1}{c}{} & \multicolumn{1}{c}{} & \multicolumn{1}{c}{Mahalanobis*} & \multicolumn{1}{c}{Gram matrix*} & \multicolumn{1}{c}{GODIN} & \multicolumn{1}{c}{Scaled Cosine} &  \multicolumn{1}{c}{SSD+} & \multicolumn{1}{c}{Mahalanobis} & \multicolumn{1}{c}{Mahalanobis*}  
        
        \\ \midrule
        
        \parbox[t]{2mm}{\multirow{16}{*}{\rotatebox[origin=c]{90}{UTKFace}}}
      
         &ID & 26 & -- & -- & -- & -- & -- & -- & --\\ 
         
         \cdashline{2-10}[1pt/1pt]
         & \multirow{15}{*}{NAS} 
         
          &    25 & 50.3{\scriptsize $\pm$0.9} & 49.9{\scriptsize $\pm$0.3} & 50.5{\scriptsize $\pm$1.8} & 51.9{\scriptsize $\pm$2.9} & 48.7{\scriptsize $\pm$1.0} & 50.1{\scriptsize $\pm$1.7} & 49.8{\scriptsize $\pm$1.0}\\ 
         & &    24 & 52.4{\scriptsize $\pm$0.8} & 50.9{\scriptsize $\pm$0.5} & 48.9{\scriptsize $\pm$1.8} & 50.9{\scriptsize $\pm$3.0} & 52.8{\scriptsize $\pm$1.2} & 53.3{\scriptsize $\pm$1.3} & 54.4{\scriptsize $\pm$1.0}\\ 
         & &    23 & 50.2{\scriptsize $\pm$1.2} & 51.7{\scriptsize $\pm$0.6} & 48.1{\scriptsize $\pm$1.3} & 49.3{\scriptsize $\pm$2.2} & 49.3{\scriptsize $\pm$1.0} & 52.7{\scriptsize $\pm$1.2} & 54.3{\scriptsize $\pm$0.5}\\ 
         & &    22 & 55.1{\scriptsize $\pm$1.2} & 56.6{\scriptsize $\pm$0.5} & 49.2{\scriptsize $\pm$1.9} & 53.6{\scriptsize $\pm$2.2} & 51.2{\scriptsize $\pm$0.7} & 56.5{\scriptsize $\pm$1.3} & 58.2{\scriptsize $\pm$0.7}\\ 
         & &    21 & 51.8{\scriptsize $\pm$1.3} & 51.3{\scriptsize $\pm$0.9} & 51.8{\scriptsize $\pm$1.3} & 51.9{\scriptsize $\pm$3.1} & 50.3{\scriptsize $\pm$0.5} & 55.3{\scriptsize $\pm$1.0} & 56.8{\scriptsize $\pm$0.8}\\ 
         & & 19-20 & 55.7{\scriptsize $\pm$1.0} & 56.6{\scriptsize $\pm$0.3} & 47.9{\scriptsize $\pm$1.2} & 54.8{\scriptsize $\pm$1.8} & 56.7{\scriptsize $\pm$0.7} & 57.9{\scriptsize $\pm$1.2} & 58.8{\scriptsize $\pm$0.3}\\ 
         & & 17-18 & 56.9{\scriptsize $\pm$1.2} & 61.3{\scriptsize $\pm$0.7} & 46.6{\scriptsize $\pm$2.0} & 58.1{\scriptsize $\pm$2.7} & 58.2{\scriptsize $\pm$1.0} & 61.0{\scriptsize $\pm$1.1} & 61.8{\scriptsize $\pm$0.6}\\ 
         & & 15-16 & 65.7{\scriptsize $\pm$1.2} & 63.3{\scriptsize $\pm$0.5} & 45.9{\scriptsize $\pm$1.9} & 66.5{\scriptsize $\pm$2.5} & 67.1{\scriptsize $\pm$1.8} & 70.7{\scriptsize $\pm$0.7} & 70.7{\scriptsize $\pm$1.5}\\ 
         & & 12-14 & 61.6{\scriptsize $\pm$1.1} & 67.7{\scriptsize $\pm$0.7} & 45.1{\scriptsize $\pm$2.8} & 69.5{\scriptsize $\pm$4.3} & 71.5{\scriptsize $\pm$1.4} & 75.3{\scriptsize $\pm$0.9} & 75.9{\scriptsize $\pm$1.9}\\
         & &  9-11 & 63.7{\scriptsize $\pm$1.0} & 70.8{\scriptsize $\pm$1.2} & 43.8{\scriptsize $\pm$2.2} & 71.3{\scriptsize $\pm$5.7} & 73.4{\scriptsize $\pm$1.6} & 80.5{\scriptsize $\pm$0.9} & 80.8{\scriptsize $\pm$1.4}\\ 
         & &   7-8 & 62.8{\scriptsize $\pm$1.4} & 73.1{\scriptsize $\pm$1.0} & 45.0{\scriptsize $\pm$1.6} & 69.4{\scriptsize $\pm$7.3} & 72.5{\scriptsize $\pm$1.4} & 82.1{\scriptsize $\pm$1.9} & 82.6{\scriptsize $\pm$1.3}\\ 
         & &   5-6 & 61.8{\scriptsize $\pm$1.3} & 70.6{\scriptsize $\pm$0.5} & 44.1{\scriptsize $\pm$3.0} & 68.9{\scriptsize $\pm$5.9} & 74.7{\scriptsize $\pm$1.4} & 83.5{\scriptsize $\pm$1.4} & 83.0{\scriptsize $\pm$1.8}\\ 
         & &   3-4 & 68.4{\scriptsize $\pm$1.0} & 75.3{\scriptsize $\pm$0.6} & 44.3{\scriptsize $\pm$1.7} & 68.1{\scriptsize $\pm$9.4} & 80.8{\scriptsize $\pm$1.3} & 88.6{\scriptsize $\pm$1.0} & 89.4{\scriptsize $\pm$1.2}\\ 
         & &     2 & 63.5{\scriptsize $\pm$1.0} & 74.4{\scriptsize $\pm$1.1} & 44.2{\scriptsize $\pm$1.9} & 65.4{\scriptsize $\pm$10.9} & 79.5{\scriptsize $\pm$1.2} & 88.3{\scriptsize $\pm$1.6} & 88.9{\scriptsize $\pm$1.3}\\ 
         & &     1 & 65.8{\scriptsize $\pm$1.9} & 80.3{\scriptsize $\pm$0.9} & 43.3{\scriptsize $\pm$2.7} & 63.2{\scriptsize $\pm$10.4} & 79.7{\scriptsize $\pm$1.8} & 90.3{\scriptsize $\pm$2.0} & 91.5{\scriptsize $\pm$1.1} \\

        \toprule

        &\multicolumn{1}{c}{\multirow{2}{*}{Distribution}} & \multicolumn{1}{c}{\multirow{2}{*}{Year}} 
        & \multicolumn{5}{c}{\multirow{1}{*}{Baselines}} & \multicolumn{2}{c}{Ours} \\ \cmidrule(lr){4-8}  \cmidrule(lr){9-10} 
        &\multicolumn{1}{c}{} & \multicolumn{1}{c}{} & \multicolumn{1}{c}{Mahalanobis*} & \multicolumn{1}{c}{Gram matrix*} & \multicolumn{1}{c}{GODIN} & \multicolumn{1}{c}{Scaled Cosine} &  \multicolumn{1}{c}{SSD+} & \multicolumn{1}{c}{Mahalanobis} & \multicolumn{1}{c}{Mahalanobis*}

        \\ \midrule
        
        \parbox[t]{2mm}{\multirow{9}{*}{\rotatebox[origin=c]{90}{Amazon Review}}}

        & ID & 2005 & -- & -- & -- & -- & -- & -- & --\\ 
        \cdashline{2-10}[1pt/1pt]
        & \multirow{9}{*}{NAS}
        
          & 2006 & 51.4{\scriptsize $\pm$0.0} & -- & 50.5{\scriptsize $\pm$0.1} & 49.4{\scriptsize $\pm$0.5} & 51.3{\scriptsize $\pm$0.3} & 51.4{\scriptsize $\pm$0.1} & 51.4{\scriptsize $\pm$0.0}\\ 
        & & 2007 & 56.6{\scriptsize $\pm$0.1} & -- & 50.9{\scriptsize $\pm$0.5} & 49.2{\scriptsize $\pm$0.6} & 55.4{\scriptsize $\pm$1.7} & 56.5{\scriptsize $\pm$0.1} & 56.6{\scriptsize $\pm$0.0}\\ 
        & & 2008 & 54.9{\scriptsize $\pm$0.0} & -- & 50.8{\scriptsize $\pm$0.1} & 49.4{\scriptsize $\pm$0.5} & 53.8{\scriptsize $\pm$0.2} & 54.9{\scriptsize $\pm$0.1} & 54.8{\scriptsize $\pm$0.1}\\ 
        & & 2009 & 53.6{\scriptsize $\pm$0.1} & -- & 50.9{\scriptsize $\pm$0.3} & 49.3{\scriptsize $\pm$0.6} & 52.5{\scriptsize $\pm$1.2} & 53.6{\scriptsize $\pm$0.1} & 53.6{\scriptsize $\pm$0.1}\\ 
        & & 2010 & 53.6{\scriptsize $\pm$0.1} & -- & 50.5{\scriptsize $\pm$0.3} & 48.8{\scriptsize $\pm$0.6} & 52.6{\scriptsize $\pm$1.3} & 53.6{\scriptsize $\pm$0.1} & 53.6{\scriptsize $\pm$0.1}\\ 
        & & 2011 & 55.0{\scriptsize $\pm$0.1} & -- & 50.5{\scriptsize $\pm$0.5} & 47.4{\scriptsize $\pm$0.7} & 53.0{\scriptsize $\pm$1.5} & 54.9{\scriptsize $\pm$0.0} & 54.8{\scriptsize $\pm$0.0}\\ 
        & & 2012 & 62.4{\scriptsize $\pm$0.1} & -- & 50.8{\scriptsize $\pm$0.6} & 45.9{\scriptsize $\pm$1.3} & 58.2{\scriptsize $\pm$2.6} & 62.6{\scriptsize $\pm$0.1} & 62.5{\scriptsize $\pm$0.1}\\ 
        & & 2013 & 75.0{\scriptsize $\pm$0.1} & -- & 51.2{\scriptsize $\pm$1.1} & 45.2{\scriptsize $\pm$1.8} & 67.4{\scriptsize $\pm$7.7} & 75.1{\scriptsize $\pm$0.0} & 75.0{\scriptsize $\pm$0.1}\\ 
        & & 2014 & 76.5{\scriptsize $\pm$0.1} & -- & 51.1{\scriptsize $\pm$0.6} & 45.8{\scriptsize $\pm$1.6} & 69.1{\scriptsize $\pm$7.2} & 76.7{\scriptsize $\pm$0.1} & 76.7{\scriptsize $\pm$0.1}\\

        \toprule

        &\multicolumn{1}{c}{\multirow{2}{*}{Distribution}} & \multicolumn{1}{c}{\multirow{2}{*}{Brightness}} 
        & \multicolumn{5}{c}{\multirow{1}{*}{Baselines}} & \multicolumn{2}{c}{Ours} \\ \cmidrule(lr){4-8}  \cmidrule(lr){9-10} 
        &\multicolumn{1}{c}{} & \multicolumn{1}{c}{} & \multicolumn{1}{c}{Mahalanobis*} & \multicolumn{1}{c}{Gram matrix*} & \multicolumn{1}{c}{GODIN} & \multicolumn{1}{c}{Scaled Cosine} &  \multicolumn{1}{c}{SSD+} & \multicolumn{1}{c}{Mahalanobis} & \multicolumn{1}{c}{Mahalanobis*}

        \\ \midrule
        
        \parbox[t]{2mm}{\multirow{16}{*}{\rotatebox[origin=c]{90}{RSNA Bone Age}}}

        &\multirow{5}{*}{NAS} & 0.0 & 100.0{\scriptsize $\pm$0.1} & 100.0{\scriptsize $\pm$0.0} & 97.8{\scriptsize $\pm$0.6} & 97.5{\scriptsize $\pm$2.1} & 100.0{\scriptsize $\pm$0.0} & 100.0{\scriptsize $\pm$0.0} & 100.0{\scriptsize $\pm$0.0}\\ 
        
         & & 0.2 & 93.9{\scriptsize $\pm$2.5} & 100.0{\scriptsize $\pm$0.0} & 48.9{\scriptsize $\pm$1.0} & 72.9{\scriptsize $\pm$9.3} & 96.1{\scriptsize $\pm$5.9} & 98.0{\scriptsize $\pm$1.0} & 89.3{\scriptsize $\pm$12.3}\\ 
         
         & & 0.4 & 77.8{\scriptsize $\pm$9.6} & 99.0{\scriptsize $\pm$0.3} & 51.8{\scriptsize $\pm$1.2} & 59.9{\scriptsize $\pm$3.4} & 92.1{\scriptsize $\pm$7.9} & 89.8{\scriptsize $\pm$2.2} & 67.2{\scriptsize $\pm$18.2}\\ 
         
         & & 0.6 & 65.2{\scriptsize $\pm$8.9} & 61.9{\scriptsize $\pm$3.5} & 50.7{\scriptsize $\pm$1.6} & 54.1{\scriptsize $\pm$3.2} & 81.9{\scriptsize $\pm$7.4} & 74.6{\scriptsize $\pm$2.5} & 58.4{\scriptsize $\pm$12.9}\\ 
         
         & & 0.8 & 54.2{\scriptsize $\pm$3.6} & 37.9{\scriptsize $\pm$1.5} & 51.1{\scriptsize $\pm$1.3} & 50.8{\scriptsize $\pm$2.2} & 62.3{\scriptsize $\pm$3.6} & 56.3{\scriptsize $\pm$1.8} & 52.0{\scriptsize $\pm$6.1}\\ 
         \cdashline{2-10}[1pt/1pt]

        & ID & 1.0 & -- & -- & -- & -- & -- & -- & --\\ 
        \cdashline{2-10}[1pt/1pt]
        
        &\multirow{8}{*}{NAS} 
        
           & 1.2 & 65.8{\scriptsize $\pm$1.8} & 58.6{\scriptsize $\pm$0.9} & 48.9{\scriptsize $\pm$1.0} & 53.5{\scriptsize $\pm$2.6} & 66.6{\scriptsize $\pm$3.2} & 55.0{\scriptsize $\pm$2.4} & 61.2{\scriptsize $\pm$1.5}\\ 
        
         & & 1.4 & 79.3{\scriptsize $\pm$1.9} & 63.5{\scriptsize $\pm$1.0} & 49.2{\scriptsize $\pm$0.6} & 58.4{\scriptsize $\pm$4.5} & 83.8{\scriptsize $\pm$5.1} & 65.1{\scriptsize $\pm$3.1} & 73.4{\scriptsize $\pm$2.6}\\ 
         
         & & 1.6 & 88.4{\scriptsize $\pm$2.4} & 69.1{\scriptsize $\pm$1.0} & 49.1{\scriptsize $\pm$1.4} & 62.9{\scriptsize $\pm$5.6} & 92.7{\scriptsize $\pm$5.4} & 75.5{\scriptsize $\pm$3.9} & 83.1{\scriptsize $\pm$5.1}\\ 
         
         & & 1.8 & 93.5{\scriptsize $\pm$1.9} & 74.9{\scriptsize $\pm$1.1} & 49.7{\scriptsize $\pm$1.4} & 66.6{\scriptsize $\pm$7.3} & 96.2{\scriptsize $\pm$4.4} & 82.3{\scriptsize $\pm$4.3} & 89.4{\scriptsize $\pm$6.0}\\ 
         
         & & 2.0 & 96.6{\scriptsize $\pm$1.1} & 80.8{\scriptsize $\pm$1.6} & 50.1{\scriptsize $\pm$0.7} & 69{\scriptsize $\pm$10.2} & 97.6{\scriptsize $\pm$3.2} & 87.0{\scriptsize $\pm$4.0} & 93.2{\scriptsize $\pm$5.4}\\ 
         
         & & 2.5 & 99.5{\scriptsize $\pm$0.2} & 91.3{\scriptsize $\pm$1.4} & 48.2{\scriptsize $\pm$2.0} & 75.4{\scriptsize $\pm$13.4} & 99.3{\scriptsize $\pm$1.1} & 93.7{\scriptsize $\pm$2.9} & 98.1{\scriptsize $\pm$2.6}\\ 
         
         & & 3.0 & 99.9{\scriptsize $\pm$0.1} & 95.7{\scriptsize $\pm$1.0} & 49.1{\scriptsize $\pm$2.7} & 83.0{\scriptsize $\pm$9.4} & 99.8{\scriptsize $\pm$0.5} & 96.6{\scriptsize $\pm$1.8} & 99.3{\scriptsize $\pm$1.0}\\ 
         
         & & 3.5 & 99.9{\scriptsize $\pm$0.1} & 97.6{\scriptsize $\pm$0.5} & 50.1{\scriptsize $\pm$3.0} & 87.3{\scriptsize $\pm$5.2} & 99.9{\scriptsize $\pm$0.2} & 97.8{\scriptsize $\pm$1.3} & 99.7{\scriptsize $\pm$0.3}\\ 
         
         & & 4.0 & 100.0{\scriptsize $\pm$0.1} & 98.4{\scriptsize $\pm$0.4} & 51.2{\scriptsize $\pm$3.5} & 89.3{\scriptsize $\pm$2.7} & 100.0{\scriptsize $\pm$0.1} & 98.4{\scriptsize $\pm$0.9} & 99.9{\scriptsize $\pm$0.1}\\ 
         
         & & 4.5 & 100.0{\scriptsize $\pm$0.0} & 98.9{\scriptsize $\pm$0.2} & 51.6{\scriptsize $\pm$3.3} & 90.2{\scriptsize $\pm$2.5} & 100.0{\scriptsize $\pm$0.1} & 98.7{\scriptsize $\pm$0.8} & 99.9{\scriptsize $\pm$0.1} \\
         \bottomrule
    \end{tabular}}
\caption{NAS detection performance on distributional shifts in three NAS datasets measured by AUROC. 
Mahalanobis and Mahalanobis* indicate using penultimate layer only or ensembling all layer respectively.
Gram matrix* also denotes the ensembled version.
}
\label{tab:recent_baseline}
\vspace{-3mm}
\end{table*}

%% file: 905metrics.tex
We report the NAS detection performance of baselines and our method based on four other metrics: Detection Accuracy, AUPR-In, AUPR-Out, and TNR@95\%TPR, which are often used in the OOD detection community.
We present the detection performance on NAS datasets in three domain measured by Detection Accuracy, AUPR-In, AUPR-Out, and TNR@95\%TPR in Tables~\ref{tab:tnr},\ref{tab:detacc},\ref{tab:auprin}, and \ref{tab:auprout} respectively. 
The results indicate that our simple modification in the training objective improves the performance of the Mahalanobis detector, thus making it a robust NAS detection method for the three NAS categories presented in the paper. 
Specifically, in UTKFace, we observe that using our suggested training loss results in a significant increase in NAS detection performance using the Mahalanobis detector. 
At the same time, the suggested method effectively detects the NAS samples in the other two datasets.
In summary, based on the results obtained using five different metrics (AUROC, Detection Accuracy, AUPR-In, AUPR-Out, and TNR@95\%TPR), our suggested modification makes the Mahalanobis detector a general NAS detection method for all three NAS categories.

\input{tabels/supp/othmetrics_tnr}
\newpage
\input{tabels/supp/othmetrics_detacc}

\newpage
\input{tabels/supp/othmetrics_auprin}
\newpage
\input{tabels/supp/othmetrics_auprout}
\newpage

%% file: tabels/supp/othmetrics_tnr.tex
\begin{table*}[h]
\small
\centering
\resizebox{\linewidth}{!}{
    \begin{tabular}{c c c  c  c  c c  c c }
        \toprule
        
        &\multicolumn{1}{c}{\multirow{2}{*}{Distribution}} & \multicolumn{1}{c}{\multirow{2}{*}{Age}} 
        & \multicolumn{5}{c}{\multirow{1}{*}{CE}}  & \multicolumn{1}{c}{Ours} \\ \cmidrule(lr){4-8}  \cmidrule(lr){9-9} 
        &\multicolumn{1}{c}{} & \multicolumn{1}{c}{} & \multicolumn{1}{c}{MSP} & \multicolumn{1}{c}{ODIN} & \multicolumn{1}{c}{Mahalanobis} & \multicolumn{1}{c}{Gram matrix} &  \multicolumn{1}{c}{Energy score} & \multicolumn{1}{c}{Mahalanobis}  
        
        \\ \midrule
        
        \parbox[t]{2mm}{\multirow{16}{*}{\rotatebox[origin=c]{90}{UTKFace}}}

         &ID & 26 & -- & -- & -- & -- & -- & -- \\
         
         \cdashline{2-9}[1pt/1pt]
         & \multirow{15}{*}{NAS} 
         
          &    25 & 6.3{\scriptsize $\pm$3.1} & 6.1{\scriptsize $\pm$1.6} & 5.5{\scriptsize $\pm$1.5} & 5.5{\scriptsize $\pm$1.8} & 6.0{\scriptsize $\pm$2.5} & 7.1{\scriptsize $\pm$2.6}\\ 
         & &    24 & 6.0{\scriptsize $\pm$1.5} & 4.1{\scriptsize $\pm$0.8} & 5.1{\scriptsize $\pm$1.7} & 6.6{\scriptsize $\pm$1.5} & 5.5{\scriptsize $\pm$2.0} & 4.6{\scriptsize $\pm$1.2}\\ 
         & &    23 & 5.3{\scriptsize $\pm$1.9} & 3.8{\scriptsize $\pm$0.9} & 4.6{\scriptsize $\pm$1.2} & 7.0{\scriptsize $\pm$1.9} & 5.4{\scriptsize $\pm$2.2} & 5.3{\scriptsize $\pm$1.5}\\  
         & &    22 & 7.2{\scriptsize $\pm$2.2} & 4.5{\scriptsize $\pm$0.9} & 5.0{\scriptsize $\pm$1.7} & 6.9{\scriptsize $\pm$2.0} & 6.7{\scriptsize $\pm$1.7} & 8.6{\scriptsize $\pm$2.6}\\  
         & &    21 & 5.7{\scriptsize $\pm$3.1} & 5.7{\scriptsize $\pm$1.2} & 6.0{\scriptsize $\pm$1.9} & 6.2{\scriptsize $\pm$1.3} & 5.9{\scriptsize $\pm$2.7} & 7.9{\scriptsize $\pm$1.8}\\  
         & & 19-20 & 6.7{\scriptsize $\pm$2.3} & 4.8{\scriptsize $\pm$1.2} & 4.8{\scriptsize $\pm$0.9} & 7.6{\scriptsize $\pm$2.5} & 6.2{\scriptsize $\pm$1.5} & 8.1{\scriptsize $\pm$2.0}\\ 
         & & 17-18 & 9.8{\scriptsize $\pm$2.4} & 7.5{\scriptsize $\pm$0.4} & 5.2{\scriptsize $\pm$1.8} & 8.9{\scriptsize $\pm$2.1} & 10.1{\scriptsize $\pm$1.9} & 11.1{\scriptsize $\pm$3.3}\\  
         & & 15-16 & 15.2{\scriptsize $\pm$5.7} & 9.4{\scriptsize $\pm$2.1} & 6.7{\scriptsize $\pm$2.1} & 11.7{\scriptsize $\pm$2.0} & 14.2{\scriptsize $\pm$4.5} & 17.1{\scriptsize $\pm$4.7}\\  
         & & 12-14 &19.2{\scriptsize $\pm$4.9} & 17.7{\scriptsize $\pm$3.5} & 4.5{\scriptsize $\pm$1.9} & 10.9{\scriptsize $\pm$3.5} & 17.2{\scriptsize $\pm$5.2} & 21.7{\scriptsize $\pm$4.7}\\ 
         & &  9-11 & 19.1{\scriptsize $\pm$6.8} & 23.9{\scriptsize $\pm$3.6} & 4.3{\scriptsize $\pm$1.7} & 10.7{\scriptsize $\pm$3.2} & 18.4{\scriptsize $\pm$5.9} & 26.5{\scriptsize $\pm$5.5}\\ 
         & &   7-8 & 17.9{\scriptsize $\pm$4.5} & 27.4{\scriptsize $\pm$5.3} & 4.9{\scriptsize $\pm$1.7} & 10.4{\scriptsize $\pm$4.3} & 18.2{\scriptsize $\pm$5.7} & 29.3{\scriptsize $\pm$7.4}\\ 
         & &   5-6 & 19.2{\scriptsize $\pm$6.0} & 28.5{\scriptsize $\pm$3.9} & 5.6{\scriptsize $\pm$0.9} & 10.0{\scriptsize $\pm$2.7} & 18.4{\scriptsize $\pm$6.0} & 29.8{\scriptsize $\pm$9.5}\\ 
         & &   3-4 & 16.3{\scriptsize $\pm$5.0} & 31.9{\scriptsize $\pm$3.1} & 6.7{\scriptsize $\pm$2.2} & 12.5{\scriptsize $\pm$3.2} & 13.5{\scriptsize $\pm$3.1} & 41.5{\scriptsize $\pm$8.8}\\ 
         & &     2 & 10.6{\scriptsize $\pm$3.5} & 37.6{\scriptsize $\pm$6.2} & 4.9{\scriptsize $\pm$1.6} & 9.0{\scriptsize $\pm$2.2} & 10.9{\scriptsize $\pm$3.3} & 39.4{\scriptsize $\pm$11.6}\\ 
         & &     1 & 12.7{\scriptsize $\pm$4.1} & 47.5{\scriptsize $\pm$2.7} & 4.9{\scriptsize $\pm$2.6} & 10.8{\scriptsize $\pm$4.1} & 10.5{\scriptsize $\pm$3.1} & 46.0{\scriptsize $\pm$15.6}\\

        \toprule

        &\multicolumn{1}{c}{\multirow{2}{*}{Distribution}} & \multicolumn{1}{c}{\multirow{2}{*}{Year}} 
        & \multicolumn{5}{c}{\multirow{1}{*}{CE}}  & \multicolumn{1}{c}{Ours} \\ \cmidrule(lr){4-8}  \cmidrule(lr){9-9} 
        &\multicolumn{1}{c}{} & \multicolumn{1}{c}{} & \multicolumn{1}{c}{MSP} & \multicolumn{1}{c}{ODIN} & \multicolumn{1}{c}{Mahalanobis} & \multicolumn{1}{c}{Gram matrix} &  \multicolumn{1}{c}{Energy score} & \multicolumn{1}{c}{Mahalanobis}

        \\ \midrule
        
        \parbox[t]{2mm}{\multirow{9}{*}{\rotatebox[origin=c]{90}{Amazon Review}}}
        
        & ID & 2005 & -- & -- & -- & -- & -- & -- \\ 
        \cdashline{2-9}[1pt/1pt]
        & \multirow{9}{*}{NAS}
           & 2006 & 5.1{\scriptsize $\pm$0.6} & 5.1{\scriptsize $\pm$0.5} & 5.4{\scriptsize $\pm$0.3} & -- & 4.8{\scriptsize $\pm$0.4} & 5.4{\scriptsize $\pm$0.2}\\ 
         
         & & 2007 & 4.8{\scriptsize $\pm$0.4} & 5.1{\scriptsize $\pm$0.5} & 8.2{\scriptsize $\pm$0.3} & -- & 4.8{\scriptsize $\pm$0.5} & 8.5{\scriptsize $\pm$0.3}\\ 
         
         & & 2008 & 4.4{\scriptsize $\pm$0.4} & 4.9{\scriptsize $\pm$0.8} & 8.1{\scriptsize $\pm$0.3} & -- & 4.5{\scriptsize $\pm$0.7} & 8.7{\scriptsize $\pm$0.2}\\ 
         
         & & 2009 & 5.0{\scriptsize $\pm$1.0} & 4.7{\scriptsize $\pm$0.5} & 7.7{\scriptsize $\pm$0.3} & -- & 4.9{\scriptsize $\pm$0.9} & 8.4{\scriptsize $\pm$0.4}\\ 
         
         & & 2010 & 4.7{\scriptsize $\pm$0.5} & 4.9{\scriptsize $\pm$0.8} & 7.8{\scriptsize $\pm$0.2} & -- & 4.8{\scriptsize $\pm$0.7} & 8.1{\scriptsize $\pm$0.2}\\  
         
         & & 2011 & 4.4{\scriptsize $\pm$0.5} & 4.4{\scriptsize $\pm$0.6} & 7.4{\scriptsize $\pm$0.4} & -- & 4.6{\scriptsize $\pm$0.7} & 8.3{\scriptsize $\pm$0.2}\\ 
         
         & & 2012 & 3.7{\scriptsize $\pm$0.4} & 4.7{\scriptsize $\pm$0.7} & 14.2{\scriptsize $\pm$0.6} & -- & 4.1{\scriptsize $\pm$1.1} & 16.4{\scriptsize $\pm$0.5}\\ 
         
         & & 2013 & 3.7{\scriptsize $\pm$0.2} & 4.5{\scriptsize $\pm$0.7} & 25.9{\scriptsize $\pm$0.8} & -- & 4.5{\scriptsize $\pm$1.5} & 30.9{\scriptsize $\pm$0.6}\\ 
         
         & & 2014 & 3.8{\scriptsize $\pm$0.5} & 5.4{\scriptsize $\pm$0.8} & 25.8{\scriptsize $\pm$0.9} & -- & 4.3{\scriptsize $\pm$1.3} & 30.2{\scriptsize $\pm$0.7}\\

        \toprule

        &\multicolumn{1}{c}{\multirow{2}{*}{Distribution}} & \multicolumn{1}{c}{\multirow{2}{*}{Brightness}} 
        & \multicolumn{5}{c}{\multirow{1}{*}{CE}}  & \multicolumn{1}{c}{Ours} \\ \cmidrule(lr){4-8}  \cmidrule(lr){9-9} 
        &\multicolumn{1}{c}{} & \multicolumn{1}{c}{} & \multicolumn{1}{c}{MSP} & \multicolumn{1}{c}{ODIN} & \multicolumn{1}{c}{Mahalanobis} & \multicolumn{1}{c}{Gram matrix} &  \multicolumn{1}{c}{Energy score} & \multicolumn{1}{c}{Mahalanobis}

        \\ \midrule

        \parbox[t]{2mm}{\multirow{16}{*}{\rotatebox[origin=c]{90}{RSNA Bone Age}}}

        &\multirow{5}{*}{NAS} & 0.0 & 40.0{\scriptsize $\pm$54.8} & 100.0{\scriptsize $\pm$0.0} & 100.0{\scriptsize $\pm$0.0} & 100.0{\scriptsize $\pm$0.0} & 40.0{\scriptsize $\pm$54.8} & 100.0{\scriptsize $\pm$0.0}\\ 
        
         & & 0.2 & 12.1{\scriptsize $\pm$4.0} & 59.3{\scriptsize $\pm$14.5} & 50.2{\scriptsize $\pm$18.8} & 17.4{\scriptsize $\pm$6.2} & 10.3{\scriptsize $\pm$6.3} & 93.6{\scriptsize $\pm$5.8}\\ 
         
         & & 0.4 & 7.2{\scriptsize $\pm$2.7} & 13.2{\scriptsize $\pm$3.3} & 15.4{\scriptsize $\pm$8.9} & 7.6{\scriptsize $\pm$2.2} & 6.1{\scriptsize $\pm$3.3} & 36.0{\scriptsize $\pm$3.7}\\ 
         
         & & 0.6 & 5.3{\scriptsize $\pm$2.0} & 4.8{\scriptsize $\pm$1.3} & 8.7{\scriptsize $\pm$4.3} & 5.1{\scriptsize $\pm$1.3} & 4.9{\scriptsize $\pm$2.6} & 14.4{\scriptsize $\pm$1.8}\\ 
         
         & & 0.8 & 5.1{\scriptsize $\pm$1.4} & 4.2{\scriptsize $\pm$1.1} & 5.9{\scriptsize $\pm$1.5} & 5.0{\scriptsize $\pm$0.8} & 4.4{\scriptsize $\pm$1.9} & 6.7{\scriptsize $\pm$1.9}\\ 
         \cdashline{2-9}[1pt/1pt]

        & ID & 1.0  & -- & -- & -- & -- & -- & -- \\  \cdashline{2-9}[1pt/1pt]

        &\multirow{8}{*}{NAS} 
        
           & 1.2 & 5.6{\scriptsize $\pm$1.9} & 8.4{\scriptsize $\pm$0.8} & 7.9{\scriptsize $\pm$1.6} & 5.7{\scriptsize $\pm$1.2} & 5.5{\scriptsize $\pm$2.1} & 6.7{\scriptsize $\pm$1.1}\\ 
        
         & & 1.4 & 7.3{\scriptsize $\pm$1.5} & 18.1{\scriptsize $\pm$3.5} & 13.0{\scriptsize $\pm$4.3} & 9.3{\scriptsize $\pm$2.5} & 7.4{\scriptsize $\pm$1.9} & 13.4{\scriptsize $\pm$2.6}\\ 
         
         & & 1.6 & 10.2{\scriptsize $\pm$3.5} & 29.3{\scriptsize $\pm$4.9} & 20.8{\scriptsize $\pm$5.9} & 13.3{\scriptsize $\pm$2.9} & 10.5{\scriptsize $\pm$3.1} & 22.8{\scriptsize $\pm$4.0}\\
         
         & & 1.8 & 11.5{\scriptsize $\pm$3.5} & 41.8{\scriptsize $\pm$7.2} & 29.0{\scriptsize $\pm$9.6} & 16.9{\scriptsize $\pm$4.0} & 10.9{\scriptsize $\pm$2.5} & 29.7{\scriptsize $\pm$7.4}\\ 
         
         & & 2.0 & 11.5{\scriptsize $\pm$3.7} & 55.2{\scriptsize $\pm$9.0} & 36.0{\scriptsize $\pm$12.2} & 18.2{\scriptsize $\pm$4.6} & 11.1{\scriptsize $\pm$2.7} & 37.8{\scriptsize $\pm$11.9}\\ 
         
         & & 2.5 & 11.5{\scriptsize $\pm$2.1} & 77.7{\scriptsize $\pm$9.6} & 53.3{\scriptsize $\pm$20.4} & 24.4{\scriptsize $\pm$6.6} & 10.4{\scriptsize $\pm$2.1} & 61.6{\scriptsize $\pm$13.9}\\ 
         
         & & 3.0 & 13.3{\scriptsize $\pm$3.1} & 92.1{\scriptsize $\pm$5.3} & 72.3{\scriptsize $\pm$23.2} & 31.7{\scriptsize $\pm$9.7} & 11.9{\scriptsize $\pm$3.2} & 82.5{\scriptsize $\pm$10.6}\\
         
         & & 3.5 & 15.5{\scriptsize $\pm$5.0} & 97.5{\scriptsize $\pm$1.9} & 81.5{\scriptsize $\pm$24.8} & 37.7{\scriptsize $\pm$10.8} & 13.8{\scriptsize $\pm$2.8} & 91.8{\scriptsize $\pm$7.6}\\ 
         
         & & 4.0 & 16.8{\scriptsize $\pm$6.1} & 99.3{\scriptsize $\pm$0.8} & 85.6{\scriptsize $\pm$22.9} & 44.0{\scriptsize $\pm$11.6} & 14.6{\scriptsize $\pm$4.1} & 96.1{\scriptsize $\pm$3.8}\\ 
         
         & & 4.5 & 18.9{\scriptsize $\pm$7.8} & 99.6{\scriptsize $\pm$0.5} & 89.9{\scriptsize $\pm$18.3} & 50.2{\scriptsize $\pm$13.8} & 16.0{\scriptsize $\pm$5.4} & 97.9{\scriptsize $\pm$2.6}\\ 
         \bottomrule
    \end{tabular}}
\caption{NAS detection performance on three NAS datasets measured by TNR at TPR 95\%. 
}
\label{tab:tnr}
\vspace{-3mm}
\end{table*}

%% file: tabels/supp/othmetrics_detacc.tex
\begin{table*}[h]
\small
\centering
\resizebox{\linewidth}{!}{
    \begin{tabular}{c c c  c  c  c c  c c }
        \toprule
        
        &\multicolumn{1}{c}{\multirow{2}{*}{Distribution}} & \multicolumn{1}{c}{\multirow{2}{*}{Age}} 
        & \multicolumn{5}{c}{\multirow{1}{*}{CE}}  & \multicolumn{1}{c}{Ours} \\ \cmidrule(lr){4-8}  \cmidrule(lr){9-9} 
        &\multicolumn{1}{c}{} & \multicolumn{1}{c}{} & \multicolumn{1}{c}{MSP} & \multicolumn{1}{c}{ODIN} & \multicolumn{1}{c}{Mahalanobis} & \multicolumn{1}{c}{Gram matrix} &  \multicolumn{1}{c}{Energy score} & \multicolumn{1}{c}{Mahalanobis}  
        
        \\ \midrule

        \parbox[t]{2mm}{\multirow{16}{*}{\rotatebox[origin=c]{90}{UTKFace}}}

         &ID & 26 & -- & --  & -- & -- & -- & -- \\ 
         
         \cdashline{2-9}[1pt/1pt]
         & \multirow{15}{*}{NAS} 
         
          &    25 & 53.1{\scriptsize $\pm$0.6} & 51.6{\scriptsize $\pm$0.8} & 53.0{\scriptsize $\pm$0.9} & 52.4{\scriptsize $\pm$0.8} & 52.8{\scriptsize $\pm$0.4} & 53.4{\scriptsize $\pm$1.7}\\  
         & &    24 & 53.9{\scriptsize $\pm$1.0} & 53.6{\scriptsize $\pm$1.0} & 53.7{\scriptsize $\pm$1.2} & 52.8{\scriptsize $\pm$0.7} & 54.0{\scriptsize $\pm$0.9} & 54.9{\scriptsize $\pm$1.0}\\ 
         & &    23 & 53.4{\scriptsize $\pm$0.5} & 52.3{\scriptsize $\pm$0.2} & 52.3{\scriptsize $\pm$1.7} & 53.0{\scriptsize $\pm$0.7} & 53.2{\scriptsize $\pm$0.9} & 54.8{\scriptsize $\pm$0.5}\\ 
         & &    22 & 53.7{\scriptsize $\pm$0.8} & 52.8{\scriptsize $\pm$1.5} & 53.4{\scriptsize $\pm$1.5} & 53.7{\scriptsize $\pm$1.4} & 54.0{\scriptsize $\pm$1.0} & 57.3{\scriptsize $\pm$1.5}\\ 
         & &    21 & 55.2{\scriptsize $\pm$1.3} & 53.8{\scriptsize $\pm$1.2} & 52.8{\scriptsize $\pm$1.8} & 53.0{\scriptsize $\pm$1.1} & 55.2{\scriptsize $\pm$1.1} & 55.6{\scriptsize $\pm$0.8}\\  
         & & 19-20 & 56.5{\scriptsize $\pm$1.3} & 53.9{\scriptsize $\pm$1.0} & 54.6{\scriptsize $\pm$1.7} & 53.2{\scriptsize $\pm$0.9} & 56.2{\scriptsize $\pm$0.9} & 57.2{\scriptsize $\pm$0.7}\\ 
         & & 17-18 & 60.8{\scriptsize $\pm$1.3} & 56.7{\scriptsize $\pm$0.5} & 55.6{\scriptsize $\pm$1.8} & 54.4{\scriptsize $\pm$0.7} & 60.2{\scriptsize $\pm$1.0} & 59.3{\scriptsize $\pm$0.6}\\ 
         & & 15-16 & 66.3{\scriptsize $\pm$1.3} & 59.5{\scriptsize $\pm$0.8} & 59.4{\scriptsize $\pm$2.0} & 56.9{\scriptsize $\pm$1.3} & 65.8{\scriptsize $\pm$1.5} & 66.4{\scriptsize $\pm$1.0}\\ 
         & & 12-14 & 70.4{\scriptsize $\pm$1.3} & 65.1{\scriptsize $\pm$1.4} & 58.2{\scriptsize $\pm$2.3} & 57.8{\scriptsize $\pm$1.8} & 70.0{\scriptsize $\pm$1.1} & 69.2{\scriptsize $\pm$0.5}\\ 
         & &  9-11 & 72.5{\scriptsize $\pm$1.4} & 71.6{\scriptsize $\pm$1.9} & 57.2{\scriptsize $\pm$2.9} & 56.2{\scriptsize $\pm$2.7} & 72.1{\scriptsize $\pm$1.4} & 74.2{\scriptsize $\pm$0.9}\\ 
         & &   7-8 & 72.8{\scriptsize $\pm$1.0} & 72.5{\scriptsize $\pm$1.6} & 55.4{\scriptsize $\pm$2.9} & 56.4{\scriptsize $\pm$1.6} & 72.6{\scriptsize $\pm$1.4} & 75.2{\scriptsize $\pm$1.6}\\ 
         & &   5-6 & 74.0{\scriptsize $\pm$1.6} & 74.8{\scriptsize $\pm$2.0} & 54.8{\scriptsize $\pm$2.2} & 55.8{\scriptsize $\pm$2.4} & 73.8{\scriptsize $\pm$1.8} & 76.2{\scriptsize $\pm$1.1}\\ 
         & &   3-4 & 75.6{\scriptsize $\pm$1.2} & 80.5{\scriptsize $\pm$2.5} & 57.2{\scriptsize $\pm$2.2} & 56.0{\scriptsize $\pm$0.7} & 74.9{\scriptsize $\pm$1.6} & 81.5{\scriptsize $\pm$0.4}\\ 
         & &     2 & 76.0{\scriptsize $\pm$1.1} & 83.3{\scriptsize $\pm$2.1} & 55.9{\scriptsize $\pm$1.0} & 54.1{\scriptsize $\pm$0.3} & 75.2{\scriptsize $\pm$1.8} & 81.6{\scriptsize $\pm$1.4}\\ 
         & &     1 & 76.9{\scriptsize $\pm$1.0} & 84.6{\scriptsize $\pm$1.2} & 54.0{\scriptsize $\pm$2.6} & 54.1{\scriptsize $\pm$2.0} & 76.2{\scriptsize $\pm$1.7} & 83.0{\scriptsize $\pm$1.7}\\

        \toprule

        &\multicolumn{1}{c}{\multirow{2}{*}{Distribution}} & \multicolumn{1}{c}{\multirow{2}{*}{Year}} 
        & \multicolumn{5}{c}{\multirow{1}{*}{CE}}  & \multicolumn{1}{c}{Ours} \\ \cmidrule(lr){4-8}  \cmidrule(lr){9-9} 
        &\multicolumn{1}{c}{} & \multicolumn{1}{c}{} & \multicolumn{1}{c}{MSP} & \multicolumn{1}{c}{ODIN} & \multicolumn{1}{c}{Mahalanobis} & \multicolumn{1}{c}{Gram matrix} &  \multicolumn{1}{c}{Energy score} & \multicolumn{1}{c}{Mahalanobis}

        \\ \midrule

        \parbox[t]{2mm}{\multirow{9}{*}{\rotatebox[origin=c]{90}{Amazon Review}}}
        
        & ID & 2005 & -- & --  & -- & -- & -- & -- \\ 
        \cdashline{2-9}[1pt/1pt]
        & \multirow{9}{*}{NAS}
           & 2006 & 50.6{\scriptsize $\pm$0.1} & 50.6{\scriptsize $\pm$0.1} & 51.7{\scriptsize $\pm$0.1} & -- & 50.5{\scriptsize $\pm$0.2} & 51.6{\scriptsize $\pm$0.1}\\ 
         
         & & 2007 & 50.2{\scriptsize $\pm$0.1} & 50.1{\scriptsize $\pm$0.2} & 54.9{\scriptsize $\pm$0.1} & -- & 50.2{\scriptsize $\pm$0.1} & 55.0{\scriptsize $\pm$0.1}\\ 
         
         & & 2008 & 50.1{\scriptsize $\pm$0.1} & 50.3{\scriptsize $\pm$0.2} & 54.4{\scriptsize $\pm$0.2} & -- & 50.2{\scriptsize $\pm$0.2} & 54.5{\scriptsize $\pm$0.2}\\ 
         
         & & 2009 & 50.4{\scriptsize $\pm$0.4} & 50.4{\scriptsize $\pm$0.2} & 53.6{\scriptsize $\pm$0.1} & -- & 50.3{\scriptsize $\pm$0.3} & 53.6{\scriptsize $\pm$0.1}\\ 
         
         & & 2010 & 50.2{\scriptsize $\pm$0.1} & 50.3{\scriptsize $\pm$0.3} & 53.3{\scriptsize $\pm$0.1} & -- & 50.2{\scriptsize $\pm$0.1} & 53.4{\scriptsize $\pm$0.1}\\
         
         & & 2011 & 50.2{\scriptsize $\pm$0.1} & 50.1{\scriptsize $\pm$0.1} & 54.5{\scriptsize $\pm$0.1} & -- & 50.3{\scriptsize $\pm$0.2} & 54.6{\scriptsize $\pm$0.1}\\ 
         
         & & 2012 & 50.0{\scriptsize $\pm$0.0} & 50.2{\scriptsize $\pm$0.2} & 60.3{\scriptsize $\pm$0.1} & -- & 50.3{\scriptsize $\pm$0.3} & 60.4{\scriptsize $\pm$0.1}\\ 
         
         & & 2013 & 50.0{\scriptsize $\pm$0.0} & 50.2{\scriptsize $\pm$0.1} & 70.3{\scriptsize $\pm$0.2} & -- & 50.4{\scriptsize $\pm$0.4} & 70.4{\scriptsize $\pm$0.2}\\ 
         
         & & 2014 & 50.1{\scriptsize $\pm$0.1} & 50.5{\scriptsize $\pm$0.3} & 70.2{\scriptsize $\pm$0.1} & -- & 50.4{\scriptsize $\pm$0.3} & 70.2{\scriptsize $\pm$0.1}\\

        \toprule

        &\multicolumn{1}{c}{\multirow{2}{*}{Distribution}} & \multicolumn{1}{c}{\multirow{2}{*}{Brightness}} 
        & \multicolumn{5}{c}{\multirow{1}{*}{CE}}  & \multicolumn{1}{c}{Ours} \\ \cmidrule(lr){4-8}  \cmidrule(lr){9-9} 
        &\multicolumn{1}{c}{} & \multicolumn{1}{c}{} & \multicolumn{1}{c}{MSP} & \multicolumn{1}{c}{ODIN} & \multicolumn{1}{c}{Mahalanobis} & \multicolumn{1}{c}{Gram matrix} &  \multicolumn{1}{c}{Energy score} & \multicolumn{1}{c}{Mahalanobis}

        \\ \midrule
        
        \parbox[t]{2mm}{\multirow{16}{*}{\rotatebox[origin=c]{90}{RSNA Bone Age}}}

        &\multirow{5}{*}{NAS} & 0.0 & 97.0{\scriptsize $\pm$2.1} & 100.0{\scriptsize $\pm$0.0} & 99.9{\scriptsize $\pm$0.1} & 99.9{\scriptsize $\pm$0.2} & 96.1{\scriptsize $\pm$2.1} & 100.0{\scriptsize $\pm$0.0}\\ 
        
         & & 0.2 & 68.3{\scriptsize $\pm$2.8} & 82.8{\scriptsize $\pm$6.3} & 89.6{\scriptsize $\pm$1.9} & 68.5{\scriptsize $\pm$3.2} & 68.0{\scriptsize $\pm$3.4} & 96.1{\scriptsize $\pm$1.2}\\ 
         
         & & 0.4 & 55.2{\scriptsize $\pm$1.2} & 62.1{\scriptsize $\pm$3.0} & 74.9{\scriptsize $\pm$4.4} & 54.8{\scriptsize $\pm$3.2} & 55.3{\scriptsize $\pm$1.6} & 84.5{\scriptsize $\pm$2.6}\\ 
         
         & & 0.6 & 52.9{\scriptsize $\pm$1.1} & 54.0{\scriptsize $\pm$1.6} & 61.9{\scriptsize $\pm$4.4} & 52.1{\scriptsize $\pm$1.7} & 53.0{\scriptsize $\pm$1.0} & 70.5{\scriptsize $\pm$3.0}\\ 
         
         & & 0.8 & 51.5{\scriptsize $\pm$0.6} & 51.1{\scriptsize $\pm$0.9} & 53.9{\scriptsize $\pm$1.7} & 51.0{\scriptsize $\pm$0.4} & 51.6{\scriptsize $\pm$0.5} & 56.0{\scriptsize $\pm$1.5}\\ 
         \cdashline{2-9}[1pt/1pt]

        & ID & 1.0 & -- & --  & -- & -- & -- & -- \\   \cdashline{2-9}[1pt/1pt]

        &\multirow{8}{*}{NAS} 
        
           & 1.2 & 53.2{\scriptsize $\pm$0.7} & 54.4{\scriptsize $\pm$0.9} & 57.3{\scriptsize $\pm$1.1} & 52.1{\scriptsize $\pm$0.9} & 53.2{\scriptsize $\pm$0.8} & 55.5{\scriptsize $\pm$2.2}\\ 
        
         & & 1.4 & 55.8{\scriptsize $\pm$0.8} & 60.4{\scriptsize $\pm$2.1} & 65.1{\scriptsize $\pm$2.2} & 55.2{\scriptsize $\pm$1.1} & 55.4{\scriptsize $\pm$0.8} & 62.5{\scriptsize $\pm$2.7}\\ 
         
         & & 1.6 & 58.6{\scriptsize $\pm$0.9} & 66.3{\scriptsize $\pm$3.0} & 71.1{\scriptsize $\pm$3.0} & 59.7{\scriptsize $\pm$1.8} & 58.7{\scriptsize $\pm$0.9} & 70.8{\scriptsize $\pm$3.0}\\ 
         
         & & 1.8 & 60.8{\scriptsize $\pm$1.1} & 73.1{\scriptsize $\pm$4.3} & 75.7{\scriptsize $\pm$4.4} & 62.4{\scriptsize $\pm$2.0} & 60.7{\scriptsize $\pm$1.1} & 76.5{\scriptsize $\pm$2.6}\\ 
         
         & & 2.0 & 61.2{\scriptsize $\pm$1.0} & 79.4{\scriptsize $\pm$4.7} & 79.2{\scriptsize $\pm$5.0} & 63.9{\scriptsize $\pm$3.5} & 61.2{\scriptsize $\pm$1.1} & 80.8{\scriptsize $\pm$3.0}\\ 
         
         & & 2.5 & 63.4{\scriptsize $\pm$3.3} & 88.5{\scriptsize $\pm$3.7} & 86.6{\scriptsize $\pm$5.2} & 68.4{\scriptsize $\pm$4.5} & 63.1{\scriptsize $\pm$2.9} & 87.7{\scriptsize $\pm$3.9}\\ 
         
         & & 3.0 & 70.4{\scriptsize $\pm$2.9} & 94.3{\scriptsize $\pm$2.5} & 91.4{\scriptsize $\pm$4.0} & 74.8{\scriptsize $\pm$3.8} & 70.3{\scriptsize $\pm$2.7} & 92.3{\scriptsize $\pm$2.8}\\
         
         & & 3.5 & 75.8{\scriptsize $\pm$1.7} & 96.9{\scriptsize $\pm$1.4} & 94.0{\scriptsize $\pm$3.4} & 78.4{\scriptsize $\pm$4.9} & 75.5{\scriptsize $\pm$1.7} & 94.8{\scriptsize $\pm$2.3}\\ 
         
         & & 4.0 & 78.9{\scriptsize $\pm$1.8} & 98.3{\scriptsize $\pm$0.9} & 95.4{\scriptsize $\pm$2.3} & 82.1{\scriptsize $\pm$5.4} & 78.6{\scriptsize $\pm$1.7} & 96.4{\scriptsize $\pm$1.7}\\ 
         
         & & 4.5 & 80.9{\scriptsize $\pm$2.5} & 98.8{\scriptsize $\pm$0.8} & 96.0{\scriptsize $\pm$2.0} & 85.0{\scriptsize $\pm$4.4} & 80.7{\scriptsize $\pm$2.4} & 97.1{\scriptsize $\pm$1.6}\\ 
         \bottomrule

    \end{tabular}}
\caption{NAS detection performance on three NAS datasets measured by detection accuracy. 
}
\label{tab:detacc}
\vspace{-3mm}
\end{table*}

%% file: tabels/supp/othmetrics_auprin.tex
\begin{table*}[h]
\small
\centering
\resizebox{\linewidth}{!}{
    \begin{tabular}{c c c  c  c  c c  c c }
        \toprule
        
        &\multicolumn{1}{c}{\multirow{2}{*}{Distribution}} & \multicolumn{1}{c}{\multirow{2}{*}{Age}} 
        & \multicolumn{5}{c}{\multirow{1}{*}{CE}}  & \multicolumn{1}{c}{Ours} \\ \cmidrule(lr){4-8}  \cmidrule(lr){9-9} 
        &\multicolumn{1}{c}{} & \multicolumn{1}{c}{} & \multicolumn{1}{c}{MSP} & \multicolumn{1}{c}{ODIN} & \multicolumn{1}{c}{Mahalanobis} & \multicolumn{1}{c}{Gram matrix} &  \multicolumn{1}{c}{Energy score} & \multicolumn{1}{c}{Mahalanobis}  
        
        \\ \midrule
        
        \parbox[t]{2mm}{\multirow{16}{*}{\rotatebox[origin=c]{90}{UTKFace}}}

         &ID & 26 &  -- & --  & -- & -- & -- & -- \\ 
         
         \cdashline{2-9}[1pt/1pt]
         & \multirow{15}{*}{NAS} 
         
          &    25 & 48.2{\scriptsize $\pm$1.4} & 46.3{\scriptsize $\pm$0.5} & 49.0{\scriptsize $\pm$0.6} & 31.9{\scriptsize $\pm$0.5} & 48.1{\scriptsize $\pm$1.3} & 48.5{\scriptsize $\pm$1.8}\\ 
         & &    24 & 51.7{\scriptsize $\pm$1.2} & 50.6{\scriptsize $\pm$0.7} & 50.1{\scriptsize $\pm$1.7} & 32.2{\scriptsize $\pm$0.4} & 51.5{\scriptsize $\pm$1.1} & 51.9{\scriptsize $\pm$1.9}\\ 
         & &    23 & 51.2{\scriptsize $\pm$1.0} & 49.7{\scriptsize $\pm$0.5} & 47.2{\scriptsize $\pm$1.2} & 32.3{\scriptsize $\pm$0.6} & 50.8{\scriptsize $\pm$0.9} & 51.2{\scriptsize $\pm$1.4}\\ 
         & &    22 & 51.7{\scriptsize $\pm$1.0} & 49.2{\scriptsize $\pm$0.7} & 49.7{\scriptsize $\pm$1.0} & 32.7{\scriptsize $\pm$0.5} & 51.6{\scriptsize $\pm$0.8} & 53.7{\scriptsize $\pm$2.4}\\ 
         & &    21 & 54.7{\scriptsize $\pm$1.4} & 53.7{\scriptsize $\pm$1.5} & 48.7{\scriptsize $\pm$1.3} & 32.4{\scriptsize $\pm$0.7} & 54.5{\scriptsize $\pm$1.4} & 53.9{\scriptsize $\pm$0.9}\\ 
         & & 19-20 & 56.3{\scriptsize $\pm$1.6} & 53.1{\scriptsize $\pm$0.7} & 51.2{\scriptsize $\pm$0.9} & 32.3{\scriptsize $\pm$0.4} & 56.0{\scriptsize $\pm$1.4} & 56.5{\scriptsize $\pm$2.4}\\ 
         & & 17-18 & 61.8{\scriptsize $\pm$1.8} & 57.5{\scriptsize $\pm$1.1} & 51.5{\scriptsize $\pm$0.9} & 33.0{\scriptsize $\pm$0.8} & 61.4{\scriptsize $\pm$1.8} & 58.5{\scriptsize $\pm$1.8}\\  
         & & 15-16 & 69.2{\scriptsize $\pm$1.3} & 63.1{\scriptsize $\pm$1.7} & 60.0{\scriptsize $\pm$1.9} & 34.3{\scriptsize $\pm$0.7} & 68.5{\scriptsize $\pm$1.8} & 70.2{\scriptsize $\pm$0.9}\\  
         & & 12-14 & 77.2{\scriptsize $\pm$1.5} & 70.5{\scriptsize $\pm$1.8} & 57.8{\scriptsize $\pm$2.9} & 34.9{\scriptsize $\pm$1.2} & 76.5{\scriptsize $\pm$1.7} & 76.4{\scriptsize $\pm$0.9}\\ 
         & &  9-11 & 79.2{\scriptsize $\pm$1.1} & 78.6{\scriptsize $\pm$2.3} & 55.1{\scriptsize $\pm$2.9} & 34.0{\scriptsize $\pm$1.6} & 78.9{\scriptsize $\pm$2.3} & 81.6{\scriptsize $\pm$1.3}\\ 
         & &   7-8 & 78.7{\scriptsize $\pm$0.8} & 80.2{\scriptsize $\pm$2.0} & 53.6{\scriptsize $\pm$2.7} & 34.1{\scriptsize $\pm$1.2} & 78.2{\scriptsize $\pm$2.0} & 83.3{\scriptsize $\pm$1.3}\\ 
         & &   5-6 & 82.1{\scriptsize $\pm$1.1} & 81.7{\scriptsize $\pm$2.4} & 52.1{\scriptsize $\pm$1.8} & 33.7{\scriptsize $\pm$1.5} & 81.9{\scriptsize $\pm$1.5} & 85.1{\scriptsize $\pm$1.0}\\ 
         & &   3-4 & 83.7{\scriptsize $\pm$1.4} & 87.4{\scriptsize $\pm$2.6} & 56.6{\scriptsize $\pm$3.3} & 33.6{\scriptsize $\pm$0.8} & 83.0{\scriptsize $\pm$2.1} & 89.9{\scriptsize $\pm$0.7}\\ 
         & &     2 & 82.7{\scriptsize $\pm$0.5} & 88.7{\scriptsize $\pm$1.6} & 54.5{\scriptsize $\pm$3.2} & 32.8{\scriptsize $\pm$0.4} & 82.3{\scriptsize $\pm$1.3} & 88.9{\scriptsize $\pm$0.9}\\ 
         & &     1 & 85.2{\scriptsize $\pm$0.6} & 92.0{\scriptsize $\pm$0.9} & 53.0{\scriptsize $\pm$4.4} & 32.5{\scriptsize $\pm$1.1} & 84.6{\scriptsize $\pm$1.4} & 91.6{\scriptsize $\pm$1.3}\\

        \toprule

        &\multicolumn{1}{c}{\multirow{2}{*}{Distribution}} & \multicolumn{1}{c}{\multirow{2}{*}{Year}} 
        & \multicolumn{5}{c}{\multirow{1}{*}{CE}}  & \multicolumn{1}{c}{Ours} \\ \cmidrule(lr){4-8}  \cmidrule(lr){9-9} 
        &\multicolumn{1}{c}{} & \multicolumn{1}{c}{} & \multicolumn{1}{c}{MSP} & \multicolumn{1}{c}{ODIN} & \multicolumn{1}{c}{Mahalanobis} & \multicolumn{1}{c}{Gram matrix} &  \multicolumn{1}{c}{Energy score} & \multicolumn{1}{c}{Mahalanobis}

        \\ \midrule
        
        \parbox[t]{2mm}{\multirow{9}{*}{\rotatebox[origin=c]{90}{Amazon Review}}}
  
        & ID & 2005 & -- & --  & -- & -- & -- & -- \\ 
        \cdashline{2-9}[1pt/1pt]
        & \multirow{9}{*}{NAS}
           & 2006 & 49.9{\scriptsize $\pm$0.2} & 49.9{\scriptsize $\pm$0.4} & 50.7{\scriptsize $\pm$0.1} & -- & 49.8{\scriptsize $\pm$0.3} & 50.9{\scriptsize $\pm$0.1}\\ 
         
         & & 2007 & 47.9{\scriptsize $\pm$0.3} & 48.4{\scriptsize $\pm$0.6} & 54.3{\scriptsize $\pm$0.1} & -- & 47.9{\scriptsize $\pm$0.4} & 54.6{\scriptsize $\pm$0.1}\\ 
         
         & & 2008 & 47.8{\scriptsize $\pm$0.3} & 48.1{\scriptsize $\pm$0.7} & 52.7{\scriptsize $\pm$0.1} & -- & 47.8{\scriptsize $\pm$0.3} & 52.9{\scriptsize $\pm$0.1}\\ 
         
         & & 2009 & 48.0{\scriptsize $\pm$0.2} & 48.5{\scriptsize $\pm$0.5} & 51.7{\scriptsize $\pm$0.1} & -- & 48.1{\scriptsize $\pm$0.3} & 51.9{\scriptsize $\pm$0.1}\\ 
         
         & & 2010 & 48.5{\scriptsize $\pm$0.2} & 49.0{\scriptsize $\pm$0.6} & 51.9{\scriptsize $\pm$0.1} & -- & 48.5{\scriptsize $\pm$0.3} & 52.2{\scriptsize $\pm$0.1}\\ 
         
         & & 2011 & 46.9{\scriptsize $\pm$0.2} & 47.7{\scriptsize $\pm$0.8} & 52.9{\scriptsize $\pm$0.1} & -- & 47.0{\scriptsize $\pm$0.2} & 53.2{\scriptsize $\pm$0.1}\\
         
         & & 2012 & 43.6{\scriptsize $\pm$0.4} & 45.1{\scriptsize $\pm$1.0} & 57.9{\scriptsize $\pm$0.1} & -- & 43.8{\scriptsize $\pm$0.4} & 58.2{\scriptsize $\pm$0.1}\\ 
         
         & & 2013 & 40.6{\scriptsize $\pm$0.3} & 42.4{\scriptsize $\pm$1.1} & 68.8{\scriptsize $\pm$0.1} & -- & 40.8{\scriptsize $\pm$0.4} & 70.3{\scriptsize $\pm$0.1}\\ 
         
         & & 2014 & 41.2{\scriptsize $\pm$0.3} & 42.7{\scriptsize $\pm$1.1} & 72.3{\scriptsize $\pm$0.3} & -- & 41.3{\scriptsize $\pm$0.4} & 74.3{\scriptsize $\pm$0.2}\\

        \toprule

        &\multicolumn{1}{c}{\multirow{2}{*}{Distribution}} & \multicolumn{1}{c}{\multirow{2}{*}{Brightness}} 
        & \multicolumn{5}{c}{\multirow{1}{*}{CE}}  & \multicolumn{1}{c}{Ours} \\ \cmidrule(lr){4-8}  \cmidrule(lr){9-9} 
        &\multicolumn{1}{c}{} & \multicolumn{1}{c}{} & \multicolumn{1}{c}{MSP} & \multicolumn{1}{c}{ODIN} & \multicolumn{1}{c}{Mahalanobis} & \multicolumn{1}{c}{Gram matrix} &  \multicolumn{1}{c}{Energy score} & \multicolumn{1}{c}{Mahalanobis}

        \\ \midrule
        
        \parbox[t]{2mm}{\multirow{16}{*}{\rotatebox[origin=c]{90}{RSNA Bone Age}}}

        &\multirow{5}{*}{NAS} & 0.0 & 96.8{\scriptsize $\pm$2.1} & 99.9{\scriptsize $\pm$0.0} & 99.8{\scriptsize $\pm$0.1} & 99.7{\scriptsize $\pm$0.3} & 95.9{\scriptsize $\pm$2.1} & 99.9{\scriptsize $\pm$0.0}\\ 
        
         & & 0.2 & 73.1{\scriptsize $\pm$4.4} & 88.8{\scriptsize $\pm$6.4} & 95.4{\scriptsize $\pm$1.3} & 43.5{\scriptsize $\pm$2.0} & 72.2{\scriptsize $\pm$5.1} & 98.5{\scriptsize $\pm$0.6}\\ 
         
         & & 0.4 & 55.7{\scriptsize $\pm$2.7} & 64.5{\scriptsize $\pm$5.4} & 81.8{\scriptsize $\pm$4.6} & 34.4{\scriptsize $\pm$2.0} & 54.8{\scriptsize $\pm$3.5} & 92.2{\scriptsize $\pm$1.9}\\ 
         
         & & 0.6 & 52.3{\scriptsize $\pm$1.7} & 53.3{\scriptsize $\pm$2.9} & 65{\scriptsize $\pm$6.4} & 32.8{\scriptsize $\pm$1.1} & 52.2{\scriptsize $\pm$1.6} & 77.4{\scriptsize $\pm$2.9}\\ 
         
         & & 0.8 & 49.9{\scriptsize $\pm$1.0} & 49.1{\scriptsize $\pm$1.2} & 53.4{\scriptsize $\pm$2.9} & 32.2{\scriptsize $\pm$0.6} & 50.0{\scriptsize $\pm$0.8} & 56.6{\scriptsize $\pm$1.6}\\ 
         \cdashline{2-9}[1pt/1pt]

        & ID & 1.0 & -- & --  & -- & -- & -- & -- \\  \cdashline{2-9}[1pt/1pt]

        &\multirow{8}{*}{NAS} 
        
           & 1.2 & 51.0{\scriptsize $\pm$1.8} & 54.2{\scriptsize $\pm$1.7} & 58.5{\scriptsize $\pm$2.6} & 32.8{\scriptsize $\pm$1.0} & 50.6{\scriptsize $\pm$1.6} & 54.8{\scriptsize $\pm$1.4}\\ 
           
         & & 1.4 & 53.0{\scriptsize $\pm$3.5} & 61.5{\scriptsize $\pm$3.3} & 69.4{\scriptsize $\pm$4.1} & 34.4{\scriptsize $\pm$0.9} & 52.2{\scriptsize $\pm$3.1} & 65.8{\scriptsize $\pm$2.3}\\ 
         
         & & 1.6 & 55.5{\scriptsize $\pm$5.5} & 69.3{\scriptsize $\pm$4.9} & 77.5{\scriptsize $\pm$4.7} & 37.0{\scriptsize $\pm$1.4} & 54.4{\scriptsize $\pm$4.7} & 77.5{\scriptsize $\pm$3.3}\\ 
         
         & & 1.8 & 58.1{\scriptsize $\pm$6.6} & 78.3{\scriptsize $\pm$5.8} & 83.0{\scriptsize $\pm$5.7} & 38.7{\scriptsize $\pm$1.6} & 56.9{\scriptsize $\pm$5.5} & 84.8{\scriptsize $\pm$3.3}\\  
         
         & & 2.0 & 60.6{\scriptsize $\pm$7.0} & 86.2{\scriptsize $\pm$5.0} & 87.0{\scriptsize $\pm$5.8} & 39.8{\scriptsize $\pm$2.5} & 59.4{\scriptsize $\pm$6.4} & 89.2{\scriptsize $\pm$3.0}\\ 
         
         & & 2.5 & 66.4{\scriptsize $\pm$7.8} & 94.9{\scriptsize $\pm$2.4} & 92.8{\scriptsize $\pm$5.1} & 43.8{\scriptsize $\pm$3.9} & 65.4{\scriptsize $\pm$8.2} & 94.8{\scriptsize $\pm$2.4}\\ 
         
         & & 3.0 & 77.3{\scriptsize $\pm$5.4} & 98.2{\scriptsize $\pm$0.9} & 96.1{\scriptsize $\pm$2.8} & 50.5{\scriptsize $\pm$4.3} & 76.6{\scriptsize $\pm$5.9} & 97.2{\scriptsize $\pm$1.5}\\ 
         
         & & 3.5 & 83.5{\scriptsize $\pm$2.8} & 99.2{\scriptsize $\pm$0.4} & 97.5{\scriptsize $\pm$2.0} & 55.7{\scriptsize $\pm$7.4} & 83.0{\scriptsize $\pm$2.9} & 98.3{\scriptsize $\pm$1.0}\\ 
         
         & & 4.0 & 87.3{\scriptsize $\pm$1.8} & 99.5{\scriptsize $\pm$0.2} & 98.2{\scriptsize $\pm$1.4} & 63.5{\scriptsize $\pm$10.9} & 86.9{\scriptsize $\pm$1.5} & 98.8{\scriptsize $\pm$0.7}\\ 
         
         & & 4.5 & 89.0{\scriptsize $\pm$2.4} & 99.6{\scriptsize $\pm$0.2} & 98.6{\scriptsize $\pm$1.1} & 69.4{\scriptsize $\pm$11.0} & 88.6{\scriptsize $\pm$2.1} & 99.0{\scriptsize $\pm$0.5}\\ 
         \bottomrule
    \end{tabular}}
\caption{NAS detection performance on three NAS datasets measured by AUPR-In. 
}
\label{tab:auprin}
\vspace{-3mm}
\end{table*}

%% file: tabels/supp/othmetrics_auprout.tex
\begin{table*}[h]
\small
\centering
\resizebox{\linewidth}{!}{
    \begin{tabular}{c c c  c  c  c c  c c }
        \toprule
        
        &\multicolumn{1}{c}{\multirow{2}{*}{Distribution}} & \multicolumn{1}{c}{\multirow{2}{*}{Age}} 
        & \multicolumn{5}{c}{\multirow{1}{*}{CE}}  & \multicolumn{1}{c}{Ours} \\ \cmidrule(lr){4-8}  \cmidrule(lr){9-9} 
        &\multicolumn{1}{c}{} & \multicolumn{1}{c}{} & \multicolumn{1}{c}{MSP} & \multicolumn{1}{c}{ODIN} & \multicolumn{1}{c}{Mahalanobis} & \multicolumn{1}{c}{Gram matrix} &  \multicolumn{1}{c}{Energy score} & \multicolumn{1}{c}{Mahalanobis}  
        
        \\ \midrule
        
        \parbox[t]{2mm}{\multirow{16}{*}{\rotatebox[origin=c]{90}{UTKFace}}}

         &ID & 26 & -- & --  & -- & -- & -- & -- \\ 
         
         \cdashline{2-9}[1pt/1pt]
         & \multirow{15}{*}{NAS} 
         
          &    25 & 51.7{\scriptsize $\pm$1.9} & 49.3{\scriptsize $\pm$1.2} & 51.2{\scriptsize $\pm$0.9} & 41.3{\scriptsize $\pm$1.2} & 51.2{\scriptsize $\pm$1.5} & 52.2{\scriptsize $\pm$2.1}\\ 
         & &    24 & 52.5{\scriptsize $\pm$1.6} & 50.1{\scriptsize $\pm$0.5} & 50.9{\scriptsize $\pm$2.7} & 42.5{\scriptsize $\pm$1.2} & 52.3{\scriptsize $\pm$1.2} & 52.9{\scriptsize $\pm$1.5}\\ 
         & &    23 & 50.9{\scriptsize $\pm$1.8} & 47.9{\scriptsize $\pm$0.5} & 49.6{\scriptsize $\pm$2.3} & 42.8{\scriptsize $\pm$1.4} & 50.2{\scriptsize $\pm$1.1} & 52.4{\scriptsize $\pm$1.7}\\  
         & &    22 & 53.4{\scriptsize $\pm$2.0} & 49.3{\scriptsize $\pm$0.9} & 51.9{\scriptsize $\pm$2.4} & 43.6{\scriptsize $\pm$1.5} & 52.9{\scriptsize $\pm$1.2} & 56.9{\scriptsize $\pm$1.1}\\  
         & &    21 & 53.5{\scriptsize $\pm$1.9} & 50.6{\scriptsize $\pm$0.6} & 51.3{\scriptsize $\pm$2.8} & 42.6{\scriptsize $\pm$1.2} & 53.0{\scriptsize $\pm$1.7} & 55.5{\scriptsize $\pm$1.4}\\  
         & & 19-20 & 55.2{\scriptsize $\pm$2.1} & 50.5{\scriptsize $\pm$0.9} & 52.7{\scriptsize $\pm$1.3} & 43.1{\scriptsize $\pm$1.2} & 54.7{\scriptsize $\pm$1.3} & 57.5{\scriptsize $\pm$1.5}\\ 
         & & 17-18 & 60.7{\scriptsize $\pm$1.9} & 55.2{\scriptsize $\pm$1.0} & 53.0{\scriptsize $\pm$1.8} & 45.4{\scriptsize $\pm$1.7} & 60.5{\scriptsize $\pm$0.9} & 60.9{\scriptsize $\pm$2.0}\\ 
         & & 15-16 & 67.8{\scriptsize $\pm$2.6} & 58.0{\scriptsize $\pm$2.0} & 56.9{\scriptsize $\pm$2.8} & 49.1{\scriptsize $\pm$1.5} & 67.2{\scriptsize $\pm$1.4} & 69.5{\scriptsize $\pm$0.5}\\ 
         & & 12-14 & 73.2{\scriptsize $\pm$2.5} & 67.2{\scriptsize $\pm$2.2} & 54.5{\scriptsize $\pm$2.4} & 49.5{\scriptsize $\pm$2.1} & 72.3{\scriptsize $\pm$2.0} & 73.4{\scriptsize $\pm$1.2}\\
         & &  9-11 & 74.2{\scriptsize $\pm$2.6} & 74.6{\scriptsize $\pm$2.8} & 53.8{\scriptsize $\pm$3.8} & 47.6{\scriptsize $\pm$4.0} & 73.6{\scriptsize $\pm$2.2} & 78.2{\scriptsize $\pm$1.8}\\ 
         & &   7-8 & 74.0{\scriptsize $\pm$2.2} & 76.1{\scriptsize $\pm$2.4} & 52.1{\scriptsize $\pm$4.5} & 48.2{\scriptsize $\pm$3.2} & 73.4{\scriptsize $\pm$2.1} & 79.9{\scriptsize $\pm$2.7}\\ 
         & &   5-6 & 75.2{\scriptsize $\pm$2.3} & 77.7{\scriptsize $\pm$3.0} & 52.7{\scriptsize $\pm$2.3} & 47.2{\scriptsize $\pm$3.2} & 74.4{\scriptsize $\pm$2.1} & 80.7{\scriptsize $\pm$3.0}\\ 
         & &   3-4 & 73.9{\scriptsize $\pm$2.4} & 81.4{\scriptsize $\pm$2.1} & 54.7{\scriptsize $\pm$3.2} & 48.1{\scriptsize $\pm$1.6} & 72.7{\scriptsize $\pm$1.6} & 86.5{\scriptsize $\pm$2.1}\\ 
         & &     2 & 71.8{\scriptsize $\pm$1.6} & 84.6{\scriptsize $\pm$2.2} & 52.1{\scriptsize $\pm$2.8} & 45.0{\scriptsize $\pm$0.9} & 71.0{\scriptsize $\pm$2.7} & 85.8{\scriptsize $\pm$3.5}\\  
         & &     1 & 73.3{\scriptsize $\pm$1.7} & 88.0{\scriptsize $\pm$1.4} & 51.1{\scriptsize $\pm$4.1} & 44.7{\scriptsize $\pm$4.0} & 71.8{\scriptsize $\pm$2.2} & 88.2{\scriptsize $\pm$3.4}\\

        \toprule

        &\multicolumn{1}{c}{\multirow{2}{*}{Distribution}} & \multicolumn{1}{c}{\multirow{2}{*}{Year}} 
        & \multicolumn{5}{c}{\multirow{1}{*}{CE}}  & \multicolumn{1}{c}{Ours} \\ \cmidrule(lr){4-8}  \cmidrule(lr){9-9} 
        &\multicolumn{1}{c}{} & \multicolumn{1}{c}{} & \multicolumn{1}{c}{MSP} & \multicolumn{1}{c}{ODIN} & \multicolumn{1}{c}{Mahalanobis} & \multicolumn{1}{c}{Gram matrix} &  \multicolumn{1}{c}{Energy score} & \multicolumn{1}{c}{Mahalanobis}

        \\ \midrule

        \parbox[t]{2mm}{\multirow{9}{*}{\rotatebox[origin=c]{90}{Amazon Review}}}
        
        & ID & 2005 & -- & --  & -- & -- & -- & -- \\ 
        \cdashline{2-9}[1pt/1pt]
        & \multirow{9}{*}{NAS}
           & 2006 & 49.8{\scriptsize $\pm$0.5} & 49.9{\scriptsize $\pm$0.3} & 51.4{\scriptsize $\pm$0.0} & -- & 49.7{\scriptsize $\pm$0.3} & 51.4{\scriptsize $\pm$0.1}\\ 
         
         & & 2007 & 48.9{\scriptsize $\pm$0.4} & 49.4{\scriptsize $\pm$0.6} & 56.0{\scriptsize $\pm$0.0} & -- & 48.8{\scriptsize $\pm$0.7} & 56.4{\scriptsize $\pm$0.1}\\ 
         
         & & 2008 & 47.9{\scriptsize $\pm$0.5} & 48.4{\scriptsize $\pm$0.7} & 54.4{\scriptsize $\pm$0.0} & -- & 48.0{\scriptsize $\pm$0.5} & 54.9{\scriptsize $\pm$0.1}\\ 
         
         & & 2009 & 48.7{\scriptsize $\pm$0.7} & 48.9{\scriptsize $\pm$0.2} & 53.7{\scriptsize $\pm$0.0} & -- & 48.6{\scriptsize $\pm$0.6} & 54.1{\scriptsize $\pm$0.0}\\ 
         
         & & 2010 & 48.2{\scriptsize $\pm$0.4} & 48.9{\scriptsize $\pm$0.8} & 53.6{\scriptsize $\pm$0.0} & -- & 48.4{\scriptsize $\pm$0.7} & 54.0{\scriptsize $\pm$0.1}\\  
         
         & & 2011 & 46.9{\scriptsize $\pm$0.4} & 47.3{\scriptsize $\pm$0.8} & 53.9{\scriptsize $\pm$0.0} & -- & 47.1{\scriptsize $\pm$0.6} & 54.5{\scriptsize $\pm$0.0}\\ 
         
         & & 2012 & 44.8{\scriptsize $\pm$0.6} & 46.8{\scriptsize $\pm$1.2} & 62.2{\scriptsize $\pm$0.0} & -- & 45.4{\scriptsize $\pm$1.2} & 63.4{\scriptsize $\pm$0.1}\\  
         
         & & 2013 & 43.6{\scriptsize $\pm$0.5} & 46.0{\scriptsize $\pm$1.0} & 73.9{\scriptsize $\pm$0.1} & -- & 44.4{\scriptsize $\pm$1.3} & 75.9{\scriptsize $\pm$0.1}\\ 
         
         & & 2014 & 44.0{\scriptsize $\pm$0.7} & 46.6{\scriptsize $\pm$1.1} & 74.4{\scriptsize $\pm$0.1} & -- & 44.5{\scriptsize $\pm$1.2} & 76.2{\scriptsize $\pm$0.1}\\

        \toprule

        &\multicolumn{1}{c}{\multirow{2}{*}{Distribution}} & \multicolumn{1}{c}{\multirow{2}{*}{Brightness}} 
        & \multicolumn{5}{c}{\multirow{1}{*}{CE}}  & \multicolumn{1}{c}{Ours} \\ \cmidrule(lr){4-8}  \cmidrule(lr){9-9} 
        &\multicolumn{1}{c}{} & \multicolumn{1}{c}{} & \multicolumn{1}{c}{MSP} & \multicolumn{1}{c}{ODIN} & \multicolumn{1}{c}{Mahalanobis} & \multicolumn{1}{c}{Gram matrix} &  \multicolumn{1}{c}{Energy score} & \multicolumn{1}{c}{Mahalanobis}

        \\ \midrule

        \parbox[t]{2mm}{\multirow{16}{*}{\rotatebox[origin=c]{90}{RSNA Bone Age}}}

        &\multirow{5}{*}{NAS} & 0.0 & 83.8{\scriptsize $\pm$8.7} & 99.9{\scriptsize $\pm$0.0} & 99.1{\scriptsize $\pm$1.1} & 99.0{\scriptsize $\pm$2.0} & 80.2{\scriptsize $\pm$7.2} & 99.9{\scriptsize $\pm$0.0}\\ 
        
         & & 0.2 & 67.2{\scriptsize $\pm$3.4} & 88.1{\scriptsize $\pm$5.7} & 89.2{\scriptsize $\pm$4.0} & 64.0{\scriptsize $\pm$2.5} & 65.7{\scriptsize $\pm$5.2} & 96.1{\scriptsize $\pm$2.4}\\ 
         
         & & 0.4 & 54.7{\scriptsize $\pm$1.7} & 62.2{\scriptsize $\pm$3.9} & 73.1{\scriptsize $\pm$5.2} & 46.4{\scriptsize $\pm$4.8} & 53.9{\scriptsize $\pm$1.7} & 84.9{\scriptsize $\pm$2.7}\\ 
         
         & & 0.6 & 52.0{\scriptsize $\pm$1.3} & 52.1{\scriptsize $\pm$2.6} & 60.3{\scriptsize $\pm$4.8} & 42.3{\scriptsize $\pm$2.9} & 51.5{\scriptsize $\pm$1.6} & 68.9{\scriptsize $\pm$1.5}\\ 
         
         & & 0.8 & 50.3{\scriptsize $\pm$0.9} & 48.9{\scriptsize $\pm$1.5} & 52.3{\scriptsize $\pm$1.4} & 40.8{\scriptsize $\pm$1.6} & 50.0{\scriptsize $\pm$1.1} & 54.3{\scriptsize $\pm$1.3}\\ 
         \cdashline{2-9}[1pt/1pt]

        & ID & 1.0 & -- & --  & -- & -- & -- & -- \\   \cdashline{2-9}[1pt/1pt]

        &\multirow{8}{*}{NAS} 
        
           & 1.2 & 51.6{\scriptsize $\pm$1.0} & 54.7{\scriptsize $\pm$1.5} & 56.5{\scriptsize $\pm$1.7} & 42.5{\scriptsize $\pm$2.1} & 51.6{\scriptsize $\pm$1.1} & 53.9{\scriptsize $\pm$2.4}\\ 
        
         & & 1.4 & 55.0{\scriptsize $\pm$1.5} & 64.4{\scriptsize $\pm$2.9} & 65.8{\scriptsize $\pm$3.5} & 47.3{\scriptsize $\pm$2.4} & 55.1{\scriptsize $\pm$1.4} & 62.9{\scriptsize $\pm$3.8}\\ 
         
         & & 1.6 & 58.9{\scriptsize $\pm$2.8} & 73.2{\scriptsize $\pm$3.6} & 73.8{\scriptsize $\pm$4.3} & 54.0{\scriptsize $\pm$2.9} & 58.9{\scriptsize $\pm$2.2} & 72.4{\scriptsize $\pm$4.9}\\ 
         
         & & 1.8 & 61.3{\scriptsize $\pm$3.2} & 81.0{\scriptsize $\pm$3.9} & 79.4{\scriptsize $\pm$5.5} & 58.1{\scriptsize $\pm$3.2} & 60.9{\scriptsize $\pm$2.2} & 78.8{\scriptsize $\pm$5.4}\\ 
         
         & & 2.0 & 61.9{\scriptsize $\pm$3.5} & 87.2{\scriptsize $\pm$3.6} & 83.2{\scriptsize $\pm$6.0} & 60.1{\scriptsize $\pm$4.7} & 61.3{\scriptsize $\pm$2.6} & 83.6{\scriptsize $\pm$5.0}\\ 
         
         & & 2.5 & 63.9{\scriptsize $\pm$3.5} & 94.7{\scriptsize $\pm$2.2} & 89.9{\scriptsize $\pm$5.6} & 66.7{\scriptsize $\pm$6.1} & 63.1{\scriptsize $\pm$3.0} & 91.6{\scriptsize $\pm$3.6}\\
         
         & & 3.0 & 69.9{\scriptsize $\pm$3.0} & 97.9{\scriptsize $\pm$1.0} & 93.8{\scriptsize $\pm$4.3} & 74.6{\scriptsize $\pm$5.5} & 69.1{\scriptsize $\pm$2.4} & 95.0{\scriptsize $\pm$2.5}\\ 
         
         & & 3.5 & 73.8{\scriptsize $\pm$3.2} & 98.9{\scriptsize $\pm$0.5} & 95.4{\scriptsize $\pm$3.8} & 79.0{\scriptsize $\pm$5.9} & 72.9{\scriptsize $\pm$2.0} & 96.5{\scriptsize $\pm$2.1}\\ 
         
         & & 4.0 & 76.6{\scriptsize $\pm$4.0} & 99.3{\scriptsize $\pm$0.4} & 96.5{\scriptsize $\pm$3.1} & 83.7{\scriptsize $\pm$6.5} & 75.4{\scriptsize $\pm$2.1} & 97.2{\scriptsize $\pm$1.8}\\ 
         
         & & 4.5 & 78.2{\scriptsize $\pm$4.7} & 99.4{\scriptsize $\pm$0.4} & 97.3{\scriptsize $\pm$2.4} & 86.9{\scriptsize $\pm$5.9} & 76.5{\scriptsize $\pm$3.5} & 97.6{\scriptsize $\pm$1.6}\\ 
         \bottomrule
    \end{tabular}}
\caption{NAS detection performance on three NAS datasets measured by AUPR-Out. 
}
\label{tab:auprout}
\vspace{-3mm}
\end{table*}